\documentclass[11pt]{article}

\usepackage[margin=1in]{geometry}
\usepackage{times}
\usepackage{graphicx}
\usepackage[round]{natbib}
\usepackage{hyperref}
\usepackage{url}
\usepackage{xcolor}
\usepackage{fancyhdr}
\usepackage{authblk}
\usepackage{amsmath}
\usepackage{amssymb}
\usepackage{booktabs}
\usepackage{multirow}
\usepackage{rotating}
\usepackage{enumitem}
\usepackage{listings}
\usepackage{float}
\usepackage{placeins}
\usepackage[most]{tcolorbox}

\definecolor{legoxpaper}{HTML}{FAFAF7}  % 直接使用页面底色
\definecolor{legoxrule}{HTML}{F4F3F0}   % 极浅边框
\renewenvironment{abstract}{%
    \begin{tcolorbox}[
        enhanced, breakable,
        colback=legoxpaper, colframe=legoxrule,
        boxrule=0.6pt, arc=7pt,
        left=16pt, right=16pt, top=11pt, bottom=11pt,
    ]%
    \small
    \begin{center}\bfseries\abstractname\end{center}
    \vspace{-0.4em}%
}{%
    \end{tcolorbox}
}

\definecolor{lstbg}{gray}{0.97}
\definecolor{lstframe}{gray}{0.75}
\definecolor{lstkey}{RGB}{20,80,160}
\definecolor{lstcomment}{RGB}{100,110,100}
\lstdefinestyle{datafmt}{
    basicstyle=\ttfamily\scriptsize,
    backgroundcolor=\color{lstbg},
    frame=single,
    rulecolor=\color{lstframe},
    framesep=5pt,
    xleftmargin=6pt,
    xrightmargin=6pt,
    breaklines=true,
    breakindent=0pt,
    breakatwhitespace=false,
    postbreak=\mbox{\textcolor{lstframe}{$\hookrightarrow$}\space},
    columns=fullflexible,
    keepspaces=true,
    showstringspaces=false,
    stringstyle=\color{lstkey},
    commentstyle=\itshape\color{lstcomment},
    aboveskip=4pt,
    belowskip=0pt,
}
\lstdefinestyle{datafmtjson}{style=datafmt,morestring=[b]"}
\lstdefinestyle{datafmttree}{style=datafmt,morestring=[b]",morecomment=[l]{\#}}

\newcommand{\legoxmark}{%
    \includegraphics[height=1.0em]{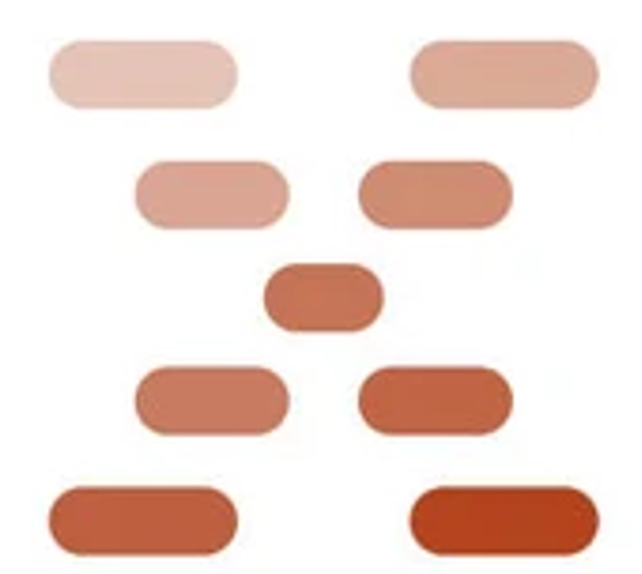}\,\textsc{LegoX}}
\fancypagestyle{plain}{%
    \fancyhf{}%
    \fancyhead[L]{\legoxmark}%
    \fancyhead[R]{\textsc{Technical Report}}%
    \fancyfoot[C]{\thepage}% 同样添加这行：确保标题页也有页码
}

\hypersetup{
    colorlinks=true,
    linkcolor=blue,
    citecolor=blue,
    urlcolor=blue
}

\title{
\textsc{Lego-RL}: Harness-Native Reinforcement Learning for \\Coding Agents
}

\author[]{
\small{
\parbox{1.0\textwidth}{
\centering
\textbf{Yiming Du}$^{1,2,*}$ \quad
\textbf{Yuxin Jiang}$^{1,*}$ \quad
\textbf{Tao Yuan}$^{1,*}$ \quad
\textbf{Jianbo Dai}$^{1}$ \quad
\textbf{Shaowei Wang}$^{1}$ \quad
\textbf{Jierun Chen}$^{1}$\\
\textbf{Chaofan Tao}$^{1}$ \quad
\textbf{Xianzhi Yu}$^{1}$ \quad
\textbf{Lifeng Shang}$^{1}$ \quad
\textbf{Kam-Fai Wong}$^{2}$ \quad
\textbf{Xiaohui Li}$^{1,\dagger}$\quad
\textbf{Haoli Bai}$^{1,\dagger}$
}
}
}

\affil{
$^{1}$Huawei Technologies Co., Ltd \quad
$^{2}$The Chinese University of Hong Kong
}

\date{}   % suppress the template's title-page date line

\makeatletter
\def\@part[#1]#2{%
    \ifnum \c@secnumdepth >\m@ne
      \refstepcounter{part}%
      \addcontentsline{toc}{part}{\thepart\hspace{1em}#1}%
    \else
      \addcontentsline{toc}{part}{#1}%
    \fi
    {\parindent \z@ \raggedright
     \interlinepenalty \@M
     \normalfont
     \ifnum \c@secnumdepth >\m@ne
       \huge\bfseries \partname\nobreakspace\thepart:\hspace{0.5em}%
     \fi
     \huge\bfseries #2\par}%
    \nobreak
    \vskip 3ex
    \@afterheading}
\makeatother

\begin{document}
\maketitle

% Unmarked title-page footnote for the author-role legend.
{\renewcommand{\thefootnote}{}\footnotetext{$^{*}$Co-first authors.\quad
$^{\dagger}$Corresponding authors: \texttt{\{lixiaohui33,baihaoli\}@huawei.com}}\setcounter{footnote}{0}}

\vspace{-3.2em}

% Release links, laid out as in the companion SWE-Lego report: one centered
% line, each address preceded by its host's mark.
\newcommand{\linkicon}[1]{\raisebox{-0.16\height}{\includegraphics[height=0.95em]{#1}}}
\begin{center}
\small
\linkicon{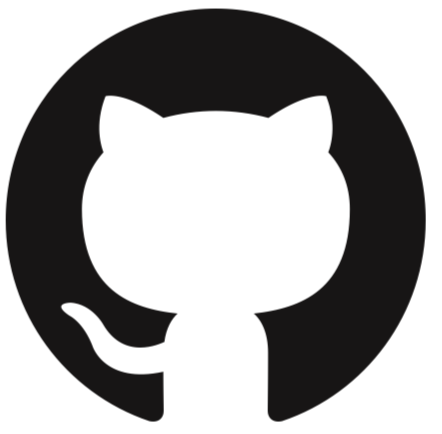}~\href{https://github.com/LegoX/Lego-RL}{\texttt{https://github.com/LegoX/Lego-RL}}\quad
\linkicon{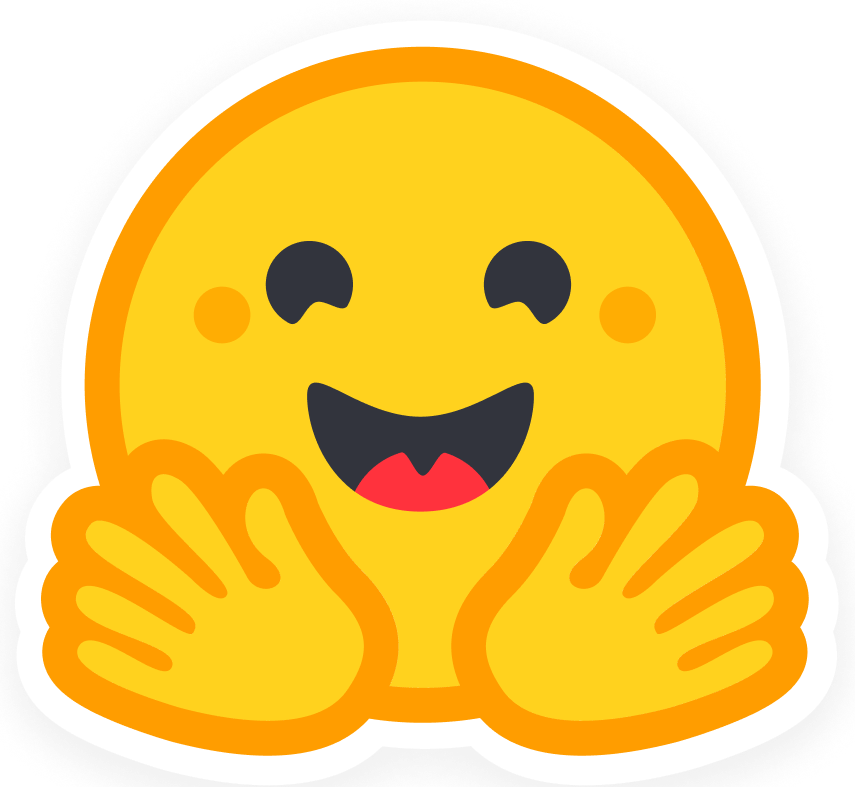}~\href{https://huggingface.co/collections/Lego-X/lego-rl}{\texttt{https://huggingface.co/LegoX/Lego-RL}}
\end{center}

\vspace{1em}

\begin{abstract}
Reinforcement learning for coding agents increasingly relies on long-running agent harnesses to manage tool integration, repository contexts, and execution feedback. However, the native execution environments of these harnesses are inherently misaligned with policy-gradient training: environmental crashes and reward hacking corrupt outcome signals, while train–inference discrepancies decouple rollout behavior from policy updates. To address this, we present \textsc{Lego-RL}, a framework that bridges native coding-agent harnesses with scalable policy-gradient optimization without modifying their internal control flow.
\textsc{Lego-RL} is built upon three pillars: \textbf{(1) faithful optimization} via in-process LLM proxying that captures raw generation streams for token-level alignment and robust trainer-side log-probability recomputation, even under harness-side compaction or re-serialization; 
% \footnote{\bai{sounds a bit hard to understand for people who first read this: harness-side history rewriting --- even when the output tokens are subtly modified by the harness?}}
\textbf{(2) reliable execution} via scalable sandbox orchestration featuring image caching and stage-wise defenses to mitigate reward hacking; and \textbf{(3) observable training} through an integrated plugin that automates validation and monitoring, paired with a Live UI for granular trajectory diagnostics.
% \textbf{Faithful}\footnote{\bai{an adj is not informative enough? consider: failthful optimization. reliable execution and observable training?}} in-process LLM proxying that captures raw generation streams for token-level alignment and robust trainer-side log-probability recomputation, even under harness-side history rewriting; \textbf{Reliable} via scalable sandbox orchestration featuring image caching and stage-wise defenses to mitigate reward hacking; and \textbf{Observable} through an integrated plugin that automates validation and monitoring, paired with a Live UI for granular trajectory diagnostics.
We evaluate \textsc{Lego-RL} by training the sparse MoE model Qwen3.5-35B-A3B with GSPO across three native coding-agent harnesses. \textsc{Lego-RL} improves Qwen3.5-35B-A3B across OpenHands SDK ($64.0\%\!\rightarrow\!70.4\%$), Claude Code ($62.4\%\!\rightarrow\!68.2\%$), and OpenCode ($57.2\%\!\rightarrow\!66.6\%$) on SWE-bench Verified, while maintaining a rollout--training probability correlation above $0.99$.
\end{abstract}

% \textsc{Lego-RL} is designed to be \textbf{Faithful:} an in-process LLM proxy that captures exact rollout token sequences at generation time for token-level alignment and faithful trainer-side log-probability recomputation despite harness-side history rewriting, extended with rollout routing replay for mixture-of-experts models; \textbf{Reliable:} scalable sandbox execution with efficient orchestration, image caching, and stage-wise defenses that protect verifier outcomes from reward hacking; and \textbf{Observable:} an agent plugin for automated pre-run validation, experiment running, training status monitoring, and real-time WebUI for trajectory-level failure diagnosis.
% We validate \textsc{Lego-RL} on production-scale workloads using a sparse MoE policy across three harnesses: Claude Code, OpenHands, and OpenCode. Coupled with the GSPO algorithm, \textsc{Lego-RL} boosts Qwen3.5-35B-A3B by $6.4$, $5.8$, and $9.4$ absolute points on SWE-bench Verified, reaching $0.682$, $0.704$, and $0.666$ respectively, while maintaining a rollout-train probability correlation exceeding $0.99$. 

\begin{figure}[t!]
    \centering
    \includegraphics[width=0.9\textwidth]{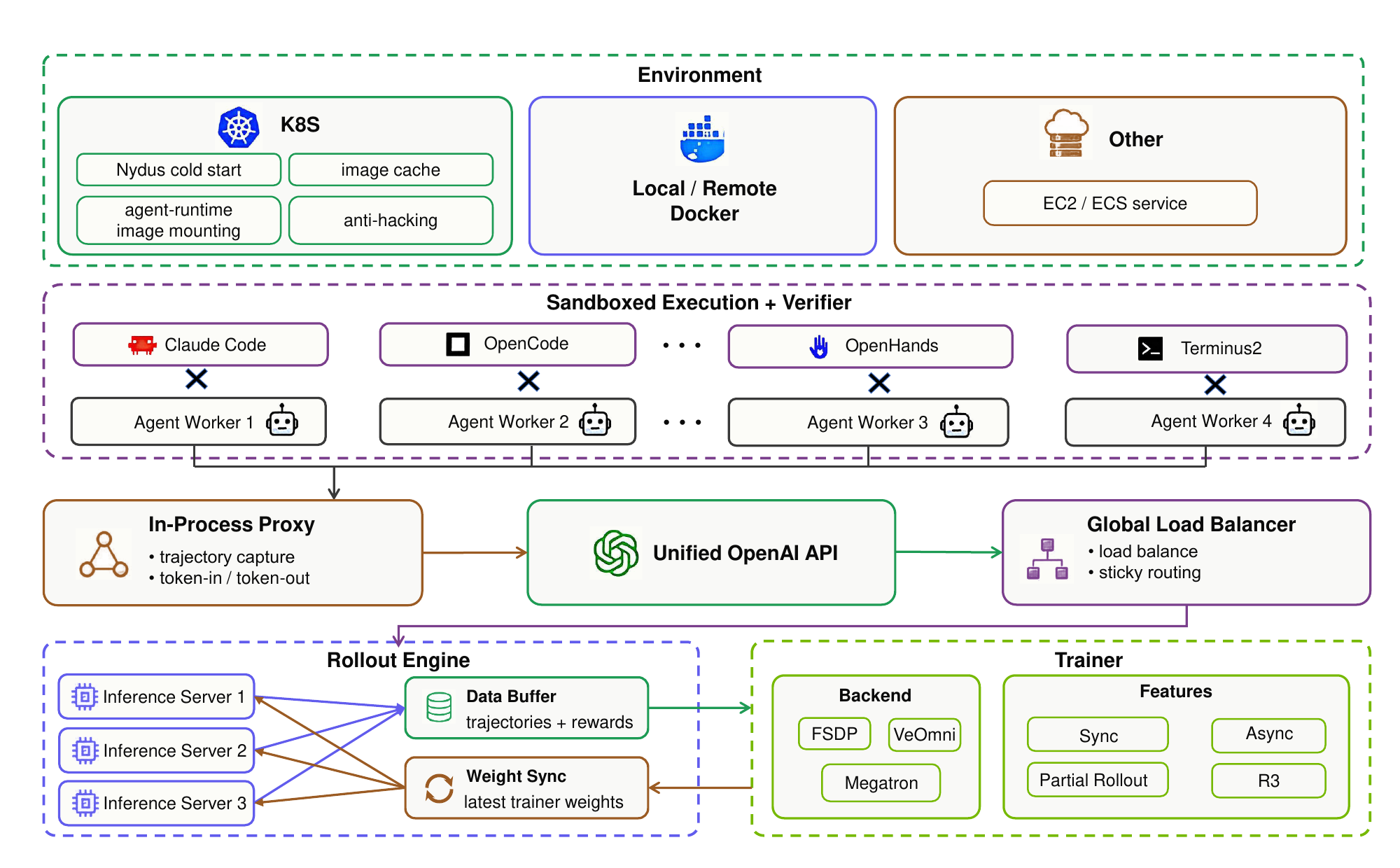}
    \caption{Overview of the \textsc{Lego-RL} training infrastructure.
    % \bai{Local/Remote Docker: put it in the middle of the box? The color scheme to be consistent with the legox colors and Figure 2? it's a bit colorful now.}
    }
    \label{fig:framework}
\end{figure}

\section{Introduction}
\label{sec:intro}

Training coding agents with reinforcement learning requires optimizing long-horizon behavior over software repositories, tools, and execution environments~\citep{wei2025swerl,golubev2025longcontext}. Rather than learning from isolated responses, reinforcement learning updates the policy from complete trajectories and their verifier rewards. A single rollout may involve repeated model calls, repository inspection, tool calling, code modification, dependency installation, and test execution before producing a sparse executable reward~\citep{jimenez2024swebench,pan2024swegym}. Because these trajectories are generated by native agent harnesses that manage prompts, tools, context, and execution state~\citep{yang2024sweagent,openhands2024}, the optimization target is the policy behavior induced by the original agent control flow.

Integrating native coding-agent harnesses such as SWE-agent~\citep{yang2024sweagent}, Claude Code~\citep{anthropic2025claudecode}, OpenHands SDK~\citep{openhands2024}, and OpenCode~\citep{opencode2025} into policy-gradient training remains challenging. Harness-side prompt construction, context compaction, and history rewriting can cause the reconstructed trajectory to differ from the exact token sequence sampled during rollout, preventing faithful trainer-side probability recomputation~\citep{xu2026polar}. Sparse mixture-of-experts models introduce an additional mismatch when rollout-time expert routing is not reproduced during training~\citep{ma2025r3}. At the execution layer, sandbox failures, dependency errors, verifier misconfiguration, timeouts, and reward hacking can discard costly trajectories or corrupt reward signals, yet are difficult to localize in asynchronous pipelines where failures propagate across stages and diagnostic signals are fragmented across workers. General RL frameworks often require coding agents to be adapted to framework-defined rollout interfaces~\citep{sheng2024hybridflow,cao2025skyrl}; for example, OpenHands-based agents may require overridden initialization, replacement of the default tool set with task-specific tools, and custom termination logic for reward extraction~\citep{cao2025skyrl, sutawika2026codescout}. Harness-native approaches preserve the existing agent workflow~\citep{yu2026openforgerl}, but scalable execution, reward-integrity protection, and trajectory-level diagnosis remain insufficiently addressed.

We present \textsc{Lego-RL}, a framework that connects existing coding-agent harnesses to scalable policy-gradient optimization without modifying their agent workflows. Built on verl~\citep{sheng2024hybridflow} and Harbor~\citep{harbor2026}, it integrates reinforcement learning and sandboxed execution around the existing harness while preserving its model APIs, tool interfaces, prompt construction, and context-management policies. Supporting a new harness requires only a lightweight adapter that launches the agent, connects it to the inference service, and returns the resulting interaction data; the remaining training pipeline is shared across agents. The framework is organized around three core pillars:
\begin{itemize}[leftmargin=1.4em,itemsep=3pt,topsep=3pt]
\item \textbf{Faithful optimization.}
An in-process LLM proxy captures exact rollout tokens and generation metadata, enabling faithful trainer-side log-probability recomputation despite harness-side history rewriting. For mixture-of-experts models, rollout-time routing decisions are replayed during training.

\item \textbf{Reliable execution.}
Scalable sandbox orchestration, image caching, and stage-wise defenses protect reward integrity, while termination-aware filtering and recovery prevent failed trajectories from disrupting asynchronous training.

\item \textbf{Observable training.}
Automated run validation, real-time monitoring, and trajectory-level diagnosis expose failures across agents, environments, verifiers, and optimization.
\end{itemize}

We evaluate \textsc{Lego-RL} across OpenHands SDK, Claude Code, and OpenCode on a sparse mixture-of-experts policy. Across the three harnesses, \textsc{Lego-RL} yields large gains in resolve rate on SWE-bench Verified: $64.0\%\!\rightarrow\!70.4\%$ on OpenHands SDK, $62.4\%\!\rightarrow\!68.2\%$ on Claude Code and $57.2\%\!\rightarrow\!66.6\%$ on OpenCode, while maintaining the rollout--training probability correlation above $0.99$. To facilitate reproducibility, we open-source the complete framework, trained models, and datasets for the community.

% of $6.4$, $5.8$, and $9.4$ percentage points on SWE-bench Verified, respectively, while maintaining a rollout--training probability correlation above $0.99$.

\section{Related Work}
\label{sec:related}

\begin{table}[!t]
    \centering
    \footnotesize
    \setlength{\tabcolsep}{3.0pt}
    \begin{tabular}{lcccccccc}
        \toprule
        & \multicolumn{4}{c}{Harness-native fidelity}
        & \multicolumn{3}{c}{Execution \& reward}
        & \multicolumn{1}{c}{Observability} \\
        \cmidrule(lr){2-5}
        \cmidrule(lr){6-8}
        \cmidrule(lr){9-9}
        
        Framework
        & \shortstack{Black-box\\harness}
        & \shortstack{Token-in/\\Token-out}
        & \shortstack{History\\alignment}
        & R3
        & \shortstack{Fully\\async}
        & \shortstack{Sandbox\\execution}
        & \shortstack{Reward-hack\\defense}
        & \shortstack{Training\\observability} \\
        
        \midrule
        
        verl~\citeyearpar{sheng2024hybridflow}
        & -- 
        & $\checkmark$
        & --
        & $\checkmark$
        & $\checkmark$
        & $\triangle$
        & --
        & $\triangle$ \\
        
        slime~\citeyearpar{thudm2025slime}
        & $\checkmark$
        & $\checkmark$
        & $\checkmark$
        & $\checkmark$
        & $\checkmark$
        & $\triangle$
        & $\triangle$
        & -- \\
        
        MOLT~\citeyearpar{nemo2026molt}
        & $\triangle$
        & $\checkmark$
        & --
        & $\checkmark$
        & $\checkmark$
        & --
        & --
        & -- \\
        
        SkyRL-Agent~\citeyearpar{cao2025skyrl}
        & $\triangle$
        & $\triangle$
        & $\triangle$
        & $\checkmark$
        & $\checkmark$
        & $\checkmark$
        & --
        & -- \\
        
        AReaL~\citeyearpar{fu2025areal}
        & $\checkmark$
        & $\triangle$
        & --
        & --
        & $\checkmark$
        & --
        & --
        & -- \\
        
        Agent Lightning~\citeyearpar{luo2025agentlightning}
        & $\triangle$
        & $\checkmark$
        & --
        & --
        & $\triangle$
        & --
        & --
        & $\triangle$ \\
        
        Polar~\citeyearpar{xu2026polar}
        & $\checkmark$
        & $\checkmark$
        & $\checkmark$
        & --
        & $\checkmark$
        & $\checkmark$
        & --
        & -- \\
        
        rLLM~\citeyearpar{rllm2026}
        & $\checkmark$
        & $\checkmark$
        & --
        & $\checkmark$
        & $\checkmark$
        & $\checkmark$
        & --
        & $\checkmark$ \\
        
        OpenForgeRL~\citeyearpar{yu2026openforgerl}
        & $\checkmark$
        & $\triangle$
        & --
        & --
        & $\checkmark$
        & $\checkmark$
        & --
        & -- \\
        
        ALE (ROLL/ROCK)~\citeyearpar{wang2025aleroll}
        & --
        & --
        & --
        & --
        & $\checkmark$
        & $\checkmark$
        & $\triangle$
        & -- \\
        
        \midrule
        
        \textsc{Lego-RL}
        & $\checkmark$
        & $\checkmark$
        & $\checkmark$
        & $\checkmark$
        & $\checkmark$
        & $\checkmark$
        & $\checkmark$
        & $\checkmark$ \\
        
        \bottomrule
    \end{tabular}
    
    \caption{Comparison of representative agentic RL frameworks.
    $\checkmark$: supported; $\triangle$: partial/conditional support;
    --: not reported. R3: rollout routing replay.
    Observability denotes monitoring training runs
    and diagnosing execution- or trajectory-level failures.}
    \label{tab:related}
\end{table}

\paragraph{Agentic RL training frameworks.}
Early LLM-RL infrastructure such as verl~\citep{sheng2024hybridflow} placed rollout generation within the trainer-managed pipeline, with the trainer directly controlling model generation and trajectory construction. Recent frameworks can be broadly categorized by where the agent rollout is implemented. slime~\citep{thudm2025slime}, MOLT~\citep{nemo2026molt}, SkyRL~\citep{cao2025skyrl}, and AReaL~\citep{fu2025areal} retain the rollout within the RL framework and provide customizable or asynchronous agentic rollout support, but existing agents must be adapted to their interaction and environment abstractions. ALE co-designs the ROLL trainer, ROCK sandbox manager, and iFlow CLI agent, achieving consistency through control of the full stack~\citep{wang2025aleroll}. In contrast, Polar, rLLM, and OpenForgeRL preserve existing agent harnesses and observe their interactions at the model API~\citep{xu2026polar,rllm2026,yu2026openforgerl}, while Agent Lightning uses SDK callbacks~\citep{luo2025agentlightning}. \textsc{Lego-RL} follows the model-API approach and integrates it with policy-gradient training, sandboxed execution, and executable verification. Table~\ref{tab:related} compares existing RL training frameworks across harness-native fidelity, execution and reward integrity, training operations, and held-out SWE-bench evaluation. Although individual capabilities are supported by several existing frameworks, \textsc{Lego-RL} brings them together in a single harness-native policy-gradient training framework.

\paragraph{Coding-agent benchmarks, harnesses, and training tasks.}
SWE-bench introduced repository-level issue resolution with executable validation~\citep{jimenez2024swebench}, later extended to harder tasks in SWE-bench Pro~\citep{deng2025swebenchpro}, multilingual repositories in Multi-SWE-bench~\citep{zan2025multiswebench}, and continuously refreshed tasks in SWE-rebench~\citep{badertdinov2025swerebench}. Unlike single-step LLM-RL, these tasks require many interleaved model calls, tool actions, and environment transitions before receiving a sparse verifier reward~\citep{zhang2025agenticrlsurvey,xi2025agentgymrl}. Harness design is therefore part of the optimization problem, since repository navigation, editing, and execution interfaces directly affect agent behavior~\citep{yang2024sweagent}. Prior work on SWE-RL either omits executable interaction~\citep{wei2025swerl} or trains within a framework-controlled agent loop~\citep{agentica2025deepswe,golubev2025longcontext}. Existing task collections provide executable environments through different construction pipelines, including SWE-Gym, R2E-Gym, SWE-Smith, OpenSWE, and SWE-Universe~\citep{pan2024swegym,jain2025r2egym,yang2025swesmith,fu2026openswe,chen2026sweuniverse}. Rather than introducing another collection, we study task selection for agentic RL based on a scalable sandbox environment, reliable verifier, and policy-relative difficulty.

\section{The \textsc{Lego-RL} Framework}
\label{sec:framework}

\subsection{Overview}
\label{sec:overview}
\textsc{Lego-RL} consists of a harness-native training infrastructure embedded within a broader closed-loop operational workflow. We first formalize the problem setup and its faithfulness requirements, then describe the training infrastructure in \S\ref{sec:training-step}, followed by the closed-loop operational workflow in \S\ref{sec:lifecycle}.

\subsubsection{Problem Setup and Objective}

\paragraph{Harness-native rollouts.}
A task instance $x=(q_x,R_x,V_x)$, drawn from a task pool $\mathcal{D}$, pairs a problem
statement $q_x$ and an initialized repository environment $R_x$ with a task-specific
executable verifier $V_x$. We treat the native coding-agent harness $\mathcal{H}$ as part of the environment and optimize only the policy $\pi_\theta$ it calls. At turn
$t=1,\dots,T$, the harness maps the current interaction and repository state $s_t$ to a
context $c_t=\mathcal{H}(s_t)$, the policy generates an assistant token span
$a_t\sim\pi_\theta(\cdot\mid c_t)$, and the harness executes the requested tool actions,
yielding $s_{t+1}$. A rollout is the sequence of prompt--response pairs actually exchanged
at the model API, $\tau=\big((c_1,a_1),\dots,(c_T,a_T)\big)$, and the verifier assigns a
single terminal binary reward $r(x,\tau)=V_x(s_{T+1})\in\{0,1\}$. Only policy-generated
tokens are trained on: let $\mathcal{M}(\tau)$ denote the set of token positions
corresponding to policy-generated response tokens. The trajectory log-likelihood is
$\log\pi_\theta(\tau)=\sum_{(t,j)\in\mathcal{M}(\tau)}\log\pi_\theta(a_{t,j}\mid c_t,a_{t,<j})$,
where each turn is conditioned on the harness-supplied context $c_t$ rather than on the raw
history---a distinction that \S\ref{sec:proxy} shows to be essential.

\paragraph{Objective.}
The expected verifier reward
$J(\theta)=\mathbb{E}_{x\sim\mathcal{D},\,\tau\sim\mathcal{H}(\pi_\theta)}\big[r(x,\tau)\big]$
is maximized with group-relative advantage estimation: each task receives a group of $G$
trajectories with rewards $r_i=r(x,\tau_i)$ and
$\hat{A}_i=(r_i-\bar{r})/(\mathrm{std}(r_{1:G})+\delta)$, where $\delta=10^{-6}$ keeps the
estimator defined when the group's reward variance is zero. Our main experiments use the GSPO~\citep{qwen2025gspo}
sequence-level surrogate
\begin{equation}
  \begin{aligned}
    \mathcal{J}_{\mathrm{GSPO}}(\theta)&=\mathbb{E}\left[\frac{1}{G}\sum_{i=1}^{G}
    w(\tau_i)\,\min\!\Big(\sigma_i(\theta)\hat{A}_i,\;
    \mathrm{clip}\big(\sigma_i(\theta),1-\epsilon_{\mathrm{low}},
    1+\epsilon_{\mathrm{high}}\big)\hat{A}_i\Big)\right],\\[0.25em]
    \sigma_i(\theta)&=\left(\frac{\pi_\theta(\tau_i)}{\pi_{\theta_{k'}}(\tau_i)}
    \right)^{1/|\mathcal{M}(\tau_i)|},
  \end{aligned}
  \label{eq:gspo}
\end{equation}
where $\theta_{k'}$ is the policy version that generated the group and
$w(\tau_i) \in \{0,1\}$ filters invalid trajectories caused by infrastructure or execution
failures, preventing them from corrupting the training signal \S\ref{sec:rollout}. The asymmetric bounds
$\epsilon_{\mathrm{low}}<\epsilon_{\mathrm{high}}$ (Table~\ref{tab:hparams}) admit more upward
than downward movement in the sequence-level ratio;
token-level PPO~\citep{schulman2017ppo} and GRPO~\citep{shao2024deepseekmath} objectives are supported by replacing $\sigma_i$ with the
per-token ratio. Two consequences shape the rest of the system: a group with equal rewards
gives $\hat{A}_i\equiv0$ and contributes no gradient, making policy-relative task difficulty
and pool composition a first-order concern, and $r$ is produced by executing code, so the
learning signal is only as trustworthy as the sandbox and the verifier.

\paragraph{Faithfulness requirement.}
The trajectory log-likelihood is well defined only if the trainer sees the exact contexts,
tokens, and mask of the rollout. Because a real harness may compact, re-serialize, or
rewrite its interaction history between turns, the recorded transcript need not decode and
re-encode back to the sampled token sequence, so $\tau$ and $\mathcal{M}(\tau)$ must be
captured at the model-serving boundary rather than reconstructed.
Let $\ell^{\mathrm{roll}}_{i,(t,j)}$ be the log-probability recorded at
generation time and $\ell^{\mathrm{train}}_{i,(t,j)}(\theta)$ the value the
trainer recomputes. Faithful optimization requires
$\ell^{\mathrm{train}}_{i,(t,j)}(\theta_{k'}) \approx
\ell^{\mathrm{roll}}_{i,(t,j)}$ for every $(t,j) \in \mathcal{M}(\tau_i)$:
agreement on the \emph{same} weights $\theta_{k'}$, up to numerical tolerance.
That holds only if token IDs, response masks, and policy weights match, and,
for sparse mixture-of-experts policies, only if training reuses the
expert-routing decisions of the behavior policy.
% Writing
% $\ell^{\mathrm{roll}}_{i,(t,j)}$ for the log-probability recorded at generation time and
% $\ell^{\mathrm{train}}_{i,(t,j)}(\theta)$ for the value recomputed by the trainer, faithful
% optimization requires
% $\ell^{\mathrm{train}}_{i,(t,j)}(\theta_{k'})\approx\ell^{\mathrm{roll}}_{i,(t,j)}$ for
% every $(t,j)\in\mathcal{M}(\tau_i)$---agreement on the \emph{same} weights $\theta_{k'}$ up
% to numerical tolerance---which demands consistent token ids, response masks, policy
% weights, and, for sparse mixture-of-experts policies, the same expert-routing decisions as
% the behavior policy.
Under fully asynchronous training the trainer version $\theta_k$ may
lead the behavior version $\theta_{k'}$; this explicitly bounded staleness is
off-policyness corrected by Eq.~\eqref{eq:gspo}, not a capture error, whereas any violation
of the agreement above is.

\subsection{Training Infrastructure}
\label{sec:training-step}

Figure \ref{fig:framework} depicts the overview of the \textsc{Lego-RL} training infrastructure. An unmodified coding-agent harness runs inside a per-trial sandbox on Kubernetes, Docker, or a cloud container service. Every model call it issues passes through the in-process proxy, which records token IDs, log-probabilities, response masks, and expert-routing decisions, and is then routed to an inference server with sticky routing. Completed trajectories and verifier rewards enter the data buffer; the trainer consumes them and pushes updated weights back to the inference servers. Only the sandbox layer is harness-specific.
% Every trial runs in a fresh, isolated sandbox that constructs the task environment, hosts the agent--repository interaction, and executes the final verifier. At scale, this layer must launch environments efficiently, isolate execution failures, and ensure that verifier rewards reflect genuine task completion. We, therefore, organize the sandbox design around environment setup and isolation, execution reliability, and reward integrity.

\subsubsection{Sandbox Execution Environment}
\label{sec:exec}

Every trial runs in a fresh, isolated sandbox, where the task environment is constructed,
the agent interacts with the repository, and the final state is evaluated through executable verification. At scale, the sandbox must support efficient environment preparation, reliable concurrent execution, and trustworthy verifier rewards. We therefore organize its design around environment setup and isolation, execution reliability, and reward integrity.

\paragraph{Environment setup and isolation.}
\textsc{Lego-RL} abstracts sandbox execution behind a common interface supporting multiple backends, such as Docker and Kubernetes. To reduce startup overhead, a Nydus lazy-pull snapshotter backed by shared storage streams image chunks on demand, avoiding full image replication across nodes. Components absent from task images, including the pinned agent runtime and grading toolchain, are mounted read-only rather than reinstalled for each trial. Tasks without prebuilt images use \emph{inline image build}, with pinned dependencies and fail-fast setup separating environment failures from policy failures. Per-pod CPU and memory limits isolate excessive resource use, while temporary package-extraction writes are redirected to an in-memory \texttt{emptyDir} to reduce local-storage contention under concurrency.

\paragraph{Execution performance and scheduling.}
Agent execution dominates trial duration, while sandbox setup and verification contribute relatively little on average but exhibit substantial tail latency. This motivates stage-specific rather than trial-wide timeouts and asynchronous scheduling, which prevents pathological trials from delaying an entire synchronized rollout batch. \S\ref{sec:efficiency} reports the measured stage-wise decomposition that supports these choices.

\paragraph{Reward integrity.}
Executable verification provides a trustworthy reward signal only when positive rewards correspond to genuine task completion. We observed both agent-side shortcuts that expose grading information and environment-side failures that make rewards independent of agent behavior. \textsc{Lego-RL} therefore enforces reward-integrity defenses within the sandbox: network restrictions are controlled by a privileged sidecar that the agent cannot modify, repository history is hidden during execution and restored when required for verification, and test dependencies are packaged into the task image to eliminate reliance on external network state. The observed failure modes, corresponding defenses, and audit results are summarized in Table~\ref{tab:antihack} of Appendix~\ref{app:antihack}.

\subsubsection{In-Process Proxy}
\label{sec:proxy}

The in-process proxy connects unmodified coding-agent harnesses to the optimizer at the provider API boundary. Co-located with the rollout engine, it supports both OpenAI-compatible and Anthropic APIs and captures token IDs, log-probabilities, response masks, and generation metadata directly from serving sessions, preserving policy-generated tokens at generation time rather than reconstructing them from the final trajectory.

\paragraph{Alignment under history rewriting.}
Because a harness may re-serialize, compact, or drop interaction history between model calls, \textsc{Lego-RL} aligns successive contexts at message granularity before assembling a training trajectory. System, user, and tool-result messages must match exactly, while tool calls are associated through their stable identifiers and function names so that argument reserialization does not alter the captured policy tokens. Matched policy-generated spans retain their original token IDs, log-probabilities, and response masks; rewritten or harness-authored content is treated only as conditioning context and is not added to $\mathcal{M}(\tau)$. Calls belonging to sub-agents are isolated from the parent capture session to prevent their tokens from entering the training trajectory. If the history cannot be aligned reliably, the affected content is excluded rather than reconstructed from the modified transcript. For sparse mixture-of-experts policies, the proxy additionally records rollout-time routing decisions and replays them during training through R3~\citep{ma2025r3}, ensuring that trainer-side probability computation follows the same expert routing as the behavior policy. \S\ref{sec:faithful} evaluates alignment under history rewriting, rollout--training agreement, and routing replay.

% Table 4 (routing-replay consistency) commented out; its numbers now run inline above.
% \begin{table}[t]
%     \centering
%     \begin{tabular}{lcccc}
%         \toprule
%         Configuration & Routing replay & Coverage & Pearson & Grad-norm \\
%         \midrule
%         MoE, replay off          & off       & $\sim\!24\%$  & $0.9946$  & $\sim\!0.36$ \\
%         MoE, replay on           & on        & $100\%$       & $0.9993$  & $\sim\!0.20$ \\
%         \bottomrule
%     \end{tabular}
%     \caption{Effect of rollout routing replay (\S\ref{sub:r3}) on train--inference
%     consistency for the sparse policy of Table~\ref{tab:main}.}
%     \label{tab:consistency}
% \end{table}

\subsubsection{Rollout and Training}
\label{sec:rollout}

Captured trials reach the optimizer through two stages that this section covers together,
because they are tuned as one system: a scheduler that keeps a fixed pool of inference slots
productive under a heavy right tail, and a verl-based trainer that consumes the filtered
batches.

\paragraph{Rollout.}
\label{sub:filter}
Agentic coding rollouts vary substantially in duration and exhibit a heavy right tail, which can reduce inference utilization under synchronized generation~\citep{fu2025areal}. \textsc{Lego-RL} therefore uses fully asynchronous rollout generation, decoupling trajectory generation from optimization and starting new rollouts as soon as previous ones finish. Configurable limits bound exceptionally long sessions, while partial trajectories spanning weight synchronization are recovered rather than discarded. Before optimization, trajectories are handled according to their termination status: execution failures are masked from training, whereas valid but incomplete policy trajectories are retained. Masked trajectories are kept for batch consistency but assigned zero optimization weight, ensuring that execution failures do not contribute to group-relative policy updates.

\paragraph{Training.}
The training layer builds on verl~\citep{sheng2024hybridflow}, inheriting support for PPO~\citep{schulman2017ppo}, GRPO~\citep{shao2024deepseekmath}, GSPO~\citep{qwen2025gspo}, vLLM~\citep{kwon2023vllm} serving, and rollout scheduling. \textsc{Lego-RL} adds the integration and backend support required for harness-native agentic RL, with \textbf{VeOmni}, \textbf{FSDP}, and \textbf{Megatron}~\citep{shoeybi2019megatron} supported as training backends. Both synchronous and fully asynchronous training are supported, with bounded policy staleness in the asynchronous setting
(\S\ref{sec:efficiency}).

\begin{figure}[t!]
    \centering
    \includegraphics[width=0.9\linewidth]{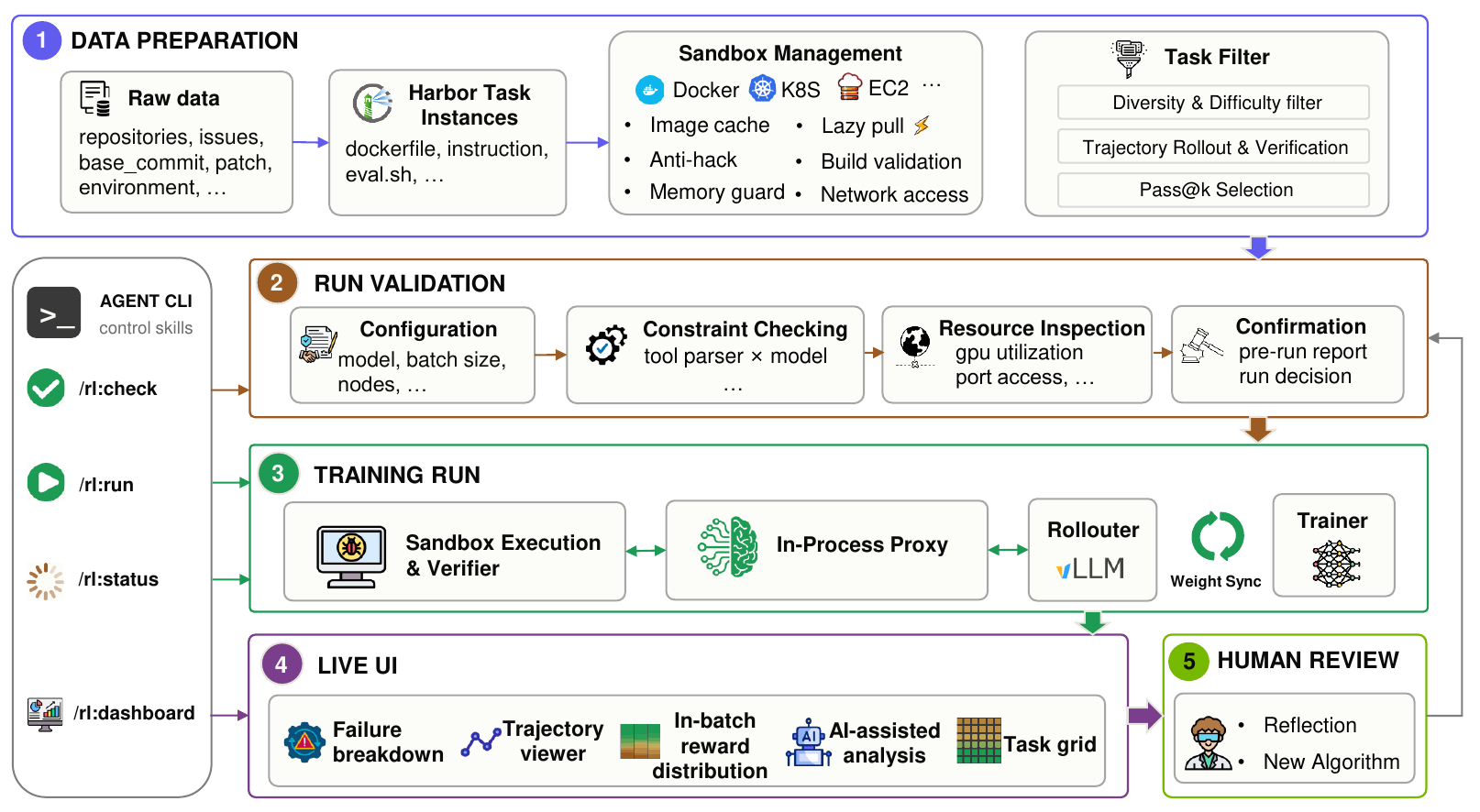}
    \caption{Closed-loop operational workflow of \textsc{Lego-RL}. The five stages cover data preparation, run validation, training, live observability, and human review. Stage~(3) corresponds to the training infrastructure shown in
    Figure~\ref{fig:framework}, while the agent plugin acts as the control plane.}
    \label{fig:pipeline}
\end{figure}

\subsection{Closed-Loop RL Workflow}
\label{sec:lifecycle}

Beyond the training infrastructure, \textsc{Lego-RL} organizes the experiment lifecycle into five stages: \textbf{Data Preparation}, \textbf{Run Validation}, \textbf{Training Run}, \textbf{Live UI}, and \textbf{Human Review} (Figure~\ref{fig:pipeline}). Stage~(1) converts task descriptions and repository snapshots into executable task instances, builds the corresponding sandbox environments, and selects training candidates based on repository diversity and policy-relative difficulty; Appendix~\ref{app:data} documents the three representations a task passes through. Stage~(2) validates experiment configurations, cross-parameter constraints, and resource availability before execution. Stage~(3) runs the sandbox-to-optimizer training loop shown in Figure~\ref{fig:framework}. Stage~(4) presents optimization metrics, rollout statistics, termination causes, and trajectory-level evidence through the Live UI. In stage~(5), researchers use this evidence to diagnose failures, analyze learned behaviors, formulate new hypotheses, and guide subsequent experiments and system refinement, thereby closing the loop. The agent plugin provides reusable skills that automate and coordinate stages~(2)--(4): \emph{Run Validation}, \emph{Training Run}, and \emph{Live Observability}. It spans these stages rather than constituting an additional workflow stage. The following subsections describe the agent plugin and Live UI in detail.

\subsubsection{Agent Plugin}
\label{sec:agentcli}

The agent plugin exposes operational capabilities through reusable skills spanning run validation, training execution, and live monitoring. Before an experiment starts, these skills resolve the experiment configuration, validate cross-parameter constraints and resource availability, and summarize the results for operator confirmation. During execution, they coordinate existing scripts and services to launch experiments, track progress, diagnose common failures, and access relevant monitoring views. Rather than duplicating training logic, the plugin composes existing system capabilities into a consistent workflow for coding agents and human operators. The same skills support training, standalone evaluation, and batch inference, reducing manual setup and configuration inconsistencies while preserving direct access to the underlying commands.

\subsubsection{Live UI}
\label{sec:dashboard}

The Live UI extends standard experiment tracking with cross-stage diagnostics for agentic RL by linking verifier outcomes to termination states, task instances, agent trajectories, tool usage, and rollout--training consistency. It provides three complementary views: termination-reason distributions and stage-level timing for identifying the source of reward variation; per-instance task grids and trajectory views for relating training metrics to task-level solve-rate changes, interaction patterns, and tool use; and consistency and in-batch distribution views for assessing rollout--training alignment and whether sampled groups retain sufficient reward variation for group-relative optimization. Together, these views connect changes in training metrics to their underlying task- and trajectory-level evidence. \S\ref{sec:behavior} evaluates this capability on representative failure cases observed during training, while Appendix~\ref{app:ui} provides the full panels, interface examples, and supporting mechanisms for
training-state tracking, verifier re-evaluation, and trajectory export.

\section{Experiments}
\label{sec:experiments}

We evaluate \textsc{Lego-RL} along five dimensions.
\S\ref{sec:setup} describes the experimental setup and \S\ref{sec:effectiveness}
reports the main results under three native coding-agent harnesses.
\S\ref{sec:data}, \S\ref{sec:faithful}, and \S\ref{sec:behavior} then evaluate
task and reward integrity, rollout--training faithfulness, and training
observability, and \S\ref{sec:efficiency} evaluates system efficiency.
% \S\ref{sec:setup} describes the  experimental setup. \S\ref{sec:effectiveness} show the mains results under three agent native harnesses. \S\ref{sec:data}, \S\ref{sec:faithful},  and \S\ref{sec:behavior} evaluate reward reliability, rollout--training faithfulness, and training observability, respectively. \S\ref{sec:efficiency} further shows the system efficiency.

\subsection{Experimental Setup}
\label{sec:setup}

\paragraph{Models and agent scaffolds.}
We use \textsc{Lego-RL} to train \textbf{Qwen3.5-35B-A3B} with the VeOmni hybrid engine under three coding-agent harnesses: OpenHands SDK~\citep{openhands2024}, Claude Code~\citep{anthropic2025claudecode}, and OpenCode~\citep{opencode2025}. Training uses group-relative advantage estimation with the GSPO sequence-level policy loss~\citep{qwen2025gspo}, implemented with verl and vLLM serving. The rollout temperature is $1.0$ and the context budget is $200$k tokens. Fully asynchronous runs use a maximum policy staleness of $1$ and recover partial rollouts across weight synchronization. Termination-aware trajectory handling is applied before optimization as described in \S\ref{sec:rollout}. Appendix~\ref{app:hparams} reports the complete resolved hyperparameters.

\paragraph{Datasets and evaluation.}
Training tasks are drawn from OpenSWE candidate pools. Unless otherwise specified, production runs use a $2{,}699$-task OpenSWE-derived index produced by the task-selection pipeline evaluated in \S\ref{sec:data}.\footnote{The training index is released at \href{https://huggingface.co/datasets/LegoX/Lego-RL-2699}{\texttt{huggingface.co/datasets/LegoX/Lego-RL-2699}}.} The training set is strictly disjoint from \emph{SWE-bench Verified} at both the repository and instance levels, preventing overlap between training and evaluation tasks. Checkpoints are evaluated on the fixed \emph{SWE-bench Verified} benchmark at temperature $0.7$ using offline executable verification, and we report solve rate as the validation score.

\begin{figure}[t]
    \centering
    \begin{minipage}{0.49\linewidth}\includegraphics[width=0.95\linewidth]{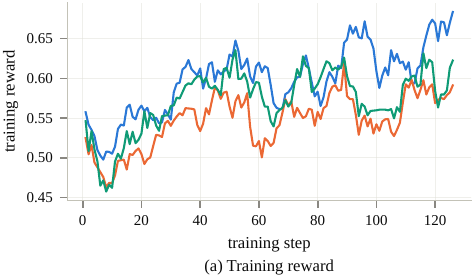}\end{minipage}\hfill
    \begin{minipage}{0.49\linewidth}\includegraphics[width=0.95\linewidth]{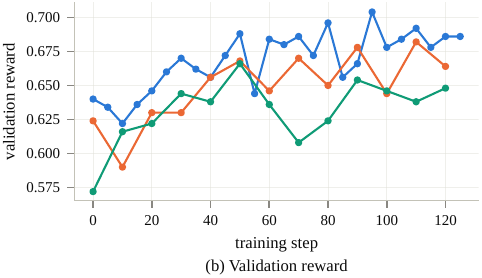}\end{minipage}\\[0.4em]
    \begin{minipage}{0.49\linewidth}\includegraphics[width=0.95\linewidth]{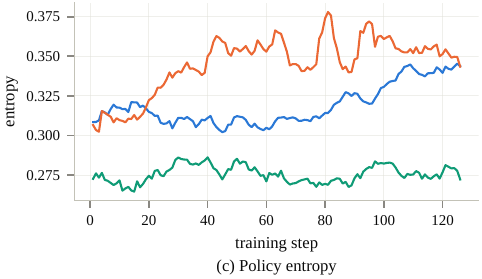}\end{minipage}\hfill
    \begin{minipage}{0.49\linewidth}\includegraphics[width=0.95\linewidth]{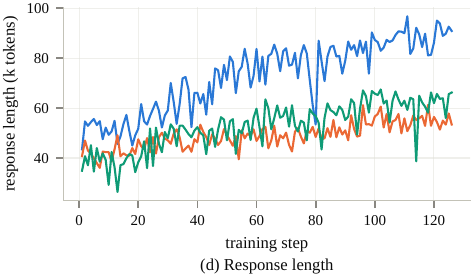}\end{minipage}\\[0.3em]
    \includegraphics{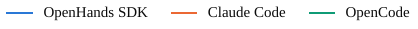}
    \caption{Training behavior of OpenHands SDK, Claude Code, and OpenCode over three epochs ($126$ training steps), showing training reward, validation reward, policy entropy, and mean response length.}
    \label{fig:scaffold}
\end{figure}

\subsection{Training Effectiveness}
\label{sec:effectiveness}
\paragraph{End-to-end effectiveness across agent scaffolds.} Figure~\ref{fig:scaffold} compares training with OpenHands SDK, Claude Code, and OpenCode using the same Qwen3.5-35B-A3B initial checkpoint, $2{,}699$-task training index, and $200$k-token context budget. Training verifier reward increases under all three scaffolds, and each run improves on SWE-bench Verified. Validation scores peak at $0.704$ with OpenHands SDK, $0.682$ with Claude Code, and $0.666$ with OpenCode, compared with step-$0$ scores of $0.640$, $0.624$, and $0.572$, corresponding to gains of $6.4$, $5.8$, and $9.4$ percentage points, respectively. Because the same initial policy yields different step-$0$ scores under different scaffolds, improvements are measured relative to each scaffold's own baseline rather than compared directly across scaffolds. Policy entropy remains stable without collapse, while mean response length increases under all three scaffolds and most strongly under OpenHands SDK, showing that the scaffold substantially affects the trajectory distribution. We further analyze the behavioral change and system efficiency in \S\ref{sec:behavior} and \S\ref{sec:efficiency}.

\paragraph{Comparison with stronger baselines.}
Table~\ref{tab:swe_bench_comparison} adds two reference points measured under the same protocol: Qwen3.6-35B-A3B, the next base generation, and KAT-Coder-V2.5-Dev~\citep{katcoder_v25_2026}, post-trained from it with supervised fine-tuning and RL. \textsc{Lego-RL}-Qwen3.5-35B-A3B is the strongest model in all three harnesses, exceeding the newer base by 3.0, 4.8, and 6.0 points---more than the 3.4, 1.0, and 3.4 points the base generation itself is worth. The gains are also harness-specific: against its own base, KAT-Coder-V2.5-Dev gains 3.4 points under Claude Code, the harness its authors report, but $-0.4$ under OpenHands SDK, where the untuned Qwen3.6-35B-A3B outscores it. We cannot isolate a cause, but the pattern is the one this work assumes: a gain obtained under one agent control flow need not survive another.

\begin{table}[t]
    \centering
    \small
    \setlength{\tabcolsep}{9pt}
    \begin{tabular}{llc}
        \toprule
        Coding agent & Model & SWE-bench Verified (\%) \\
        \midrule
        \multirow{4}{*}{OpenHands SDK}
            & Qwen3.5-35B-A3B~\citep{qwen3.5}         & $64.0$ \\
            & Qwen3.6-35B-A3B~\citep{qwen36_35b_a3b}         & $67.4$ \\
            & KAT-Coder-V2.5-Dev~\citep{katcoder_v25_2026}             & $67.0$ \\
            & \textsc{Lego-RL}-Qwen3.5-35B-A3B & $\mathbf{70.4\ (+6.4)}$ \\
        \cmidrule(lr){1-3}
        \multirow{4}{*}{Claude Code}
            & Qwen3.5-35B-A3B~\citep{qwen3.5}         & $62.4$ \\
            & Qwen3.6-35B-A3B~\citep{qwen36_35b_a3b}         & $63.4$ \\
            & KAT-Coder-V2.5-Dev~\citep{katcoder_v25_2026}       & $66.8$ \\
            & \textsc{Lego-RL}-Qwen3.5-35B-A3B & $\mathbf{68.2\ (+5.8)}$ \\
        \cmidrule(lr){1-3}
        \multirow{4}{*}{OpenCode}
            & Qwen3.5-35B-A3B~\citep{qwen3.5}         & $57.2$ \\
            & Qwen3.6-35B-A3B~\citep{qwen36_35b_a3b}         & $60.6$ \\
            & KAT-Coder-V2.5-Dev~\citep{katcoder_v25_2026}             & $64.8$ \\
            & \textsc{Lego-RL}-Qwen3.5-35B-A3B & $\mathbf{66.6\ (+9.4)}$ \\
        \bottomrule
    \end{tabular}
    \caption{SWE-bench Verified performance across the three coding agents.
All numbers are measured by us under the same harness version and evaluation
protocol (temperature 0.7, 200 turns, 200k context budget).}
    \label{tab:swe_bench_comparison}
\end{table}

\subsection{Task Reliability and Reward Integrity}
\label{sec:data}

Reliable RL depends on three properties: \emph{task validity}, ensuring that tasks can be executed and graded correctly; \emph{trajectory validity}, ensuring that rollout outcomes reflect policy behavior rather than infrastructure failures; and \emph{reward informativeness}, ensuring sufficient within-group reward variation for effective policy optimization. \textsc{Lego-RL} addresses these properties through task screening, termination-aware trajectory admission, and difficulty-aware task selection.

\paragraph{Task validity.}
\textsc{Lego-RL} first applies static filtering to the $36{,}884$ OpenSWE-derived candidates. Rule-based screening enforces basic validity, repository diversity, and coarse complexity constraints, reducing the pool to $22{,}806$ tasks. Build and verifier validation further remove tasks that cannot be executed or graded reliably, leaving $21{,}681$ tasks; approximately $2.5\%$ of inspected tasks contain verifier logic that incorrectly applies the reference patch. The remaining tasks undergo rollout-based difficulty screening using Qwen3.6-27B with the OpenHands SDK scaffold. Retaining tasks solved $1$--$3$ times in four trials produces the final $2{,}699$-task training index. Although screening uses a single model--scaffold configuration, the resulting task pool also supports effective training with Claude Code and OpenCode.

\begin{figure}[!t]
    \centering
    \includegraphics[width=\linewidth]{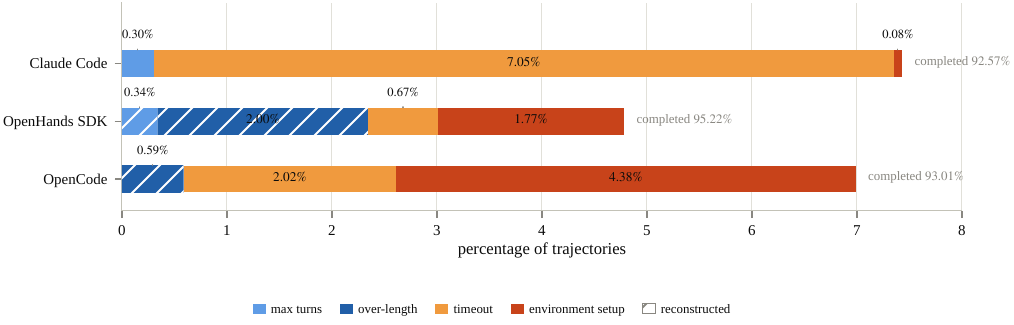}
    \caption{Trajectory termination profiles across agent scaffolds. Bars show the proportion of trajectories by termination reason; timeout and environment-setup failures are excluded from optimization.}
    \label{fig:admission}
\end{figure}

\paragraph{Trajectory validity.}
Task-level validation cannot guarantee that every rollout produces a trustworthy training outcome. \textsc{Lego-RL} therefore applies termination-aware admission before optimization: trajectories ending in infrastructure failures are excluded from group-relative advantage estimation and the policy loss, whereas valid trajectories reaching configured turn or token limits retain their verifier outcomes. Figure~\ref{fig:admission} shows that $7.1\%$ of Claude Code, $2.4\%$ of OpenHands SDK, and $6.4\%$ of OpenCode trajectories are excluded from optimization. Termination profiles differ across scaffolds, with wall-clock timeouts dominating under Claude Code and environment-setup failures under OpenCode. These differences come from the harness as much as from the infrastructure: the same sandbox stack produces a different termination mix under each harness.

\begin{figure}[!t]
    \centering
    \includegraphics[width=0.9\linewidth]{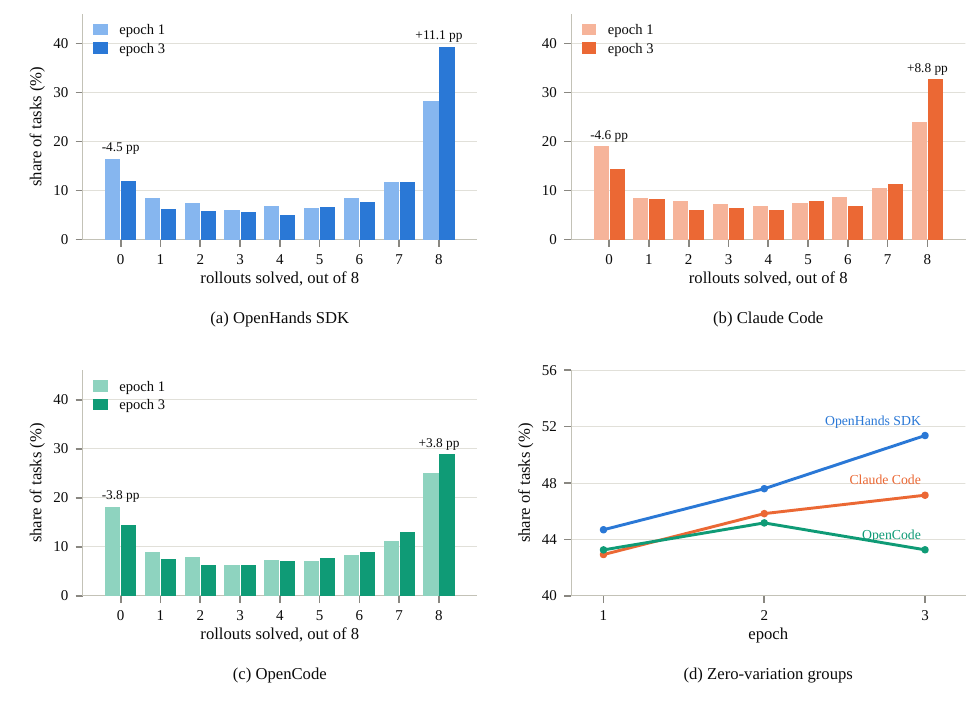}
    \caption{In-batch reward distributions across training. \textbf{(a--c)}~Distribution of tasks by the number of successful rollouts out of eight in the first and last epoch for each scaffold. The $0/8$ and $8/8$ groups provide no group-relative advantage. \textbf{(d)}~Combined proportion of these zero-variation groups across epochs.}
    \label{fig:rewarddist}
\end{figure}

\paragraph{Reward informativeness.} With eight rollouts per task, groups solved either zero or eight times provide no within-group reward variation and therefore no group-relative learning signal. Comparing the first and last epoch (Figure~\ref{fig:rewarddist}), the proportion of all-wrong groups decreases while the proportion of all-solved groups increases across all scaffolds, showing that the informativeness of a fixed task pool changes as the policy improves. Under OpenHands SDK, the proportion of zero-variation groups increases from $44.7\%$ to $51.4\%$, as the growth in all-solved groups outweighs the decline in all-wrong groups. Under OpenCode, the smaller increase in all-solved groups leaves the zero-variation proportion approximately unchanged at $43.3\%$. These results show that task difficulty is policy-relative and that a fixed task pool can gradually provide less informative group-relative supervision.

% \paragraph{Task difficulty shifts during training.}
% With eight rollouts per task, groups solved either zero or eight times provide no within-group reward variation and therefore no group-relative learning signal. Comparing the first and last epoch (Figure~\ref{fig:rewarddist}), the proportion of all-wrong groups decreases while the proportion of all-solved groups increases under every scaffold, indicating that the effective difficulty of a fixed task pool changes as the policy improves. Under OpenHands, the proportion of groups with no reward variation increases from $44.7\%$ to $51.4\%$, as repeated success improves faster than task coverage. OpenCode shows a smaller increase in the proportion of all-solved groups, leaving the proportion of groups with no reward variation approximately unchanged at $43.3\%$. This is consistent with its weaker and less sustained improvement during later training
% (Figure~\ref{fig:scaffold}). These results show that task difficulty is policy-relative and that a fixed training pool can gradually provide less informative group-relative supervision.

\FloatBarrier

\begin{figure}[!t]
    \centering
    \begin{minipage}{0.49\linewidth}\includegraphics[width=\linewidth]{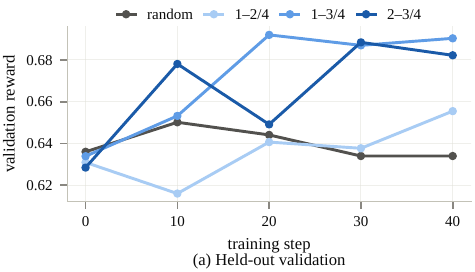}\end{minipage}\hfill
    \begin{minipage}{0.49\linewidth}\includegraphics[width=\linewidth]{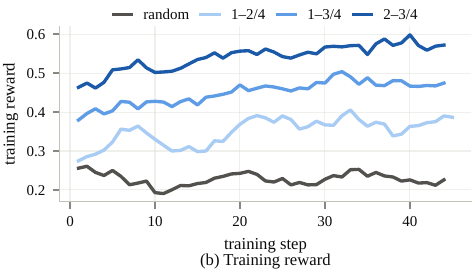}\end{minipage}
    \caption{Task-selection ablation across four $951$-task pools. \textbf{(a)}~Held-out validation solve rate, measured over the validation tasks that executed. \textbf{(b)}~Training verifier reward; levels are pool-specific, so only the slopes are comparable.}
    \label{fig:taskband}
\end{figure}

\paragraph{Effect of difficulty screening.}
We further isolate the role of difficulty screening by comparing four $951$-task training pools under otherwise matched configurations: the lower and upper halves of the selected difficulty band, the full band, and a random sample from the unscreened pool. A validation task whose rollout is lost to a harness failure is scored zero, not retried, and the four arms ran at different inference concurrency. We therefore report solve rates over the tasks that executed. This correction
moves an arm by 0.6 to 5.2 percentage points, and it is what makes the arms comparable: all four initial passes evaluate the same checkpoint, and once corrected they agree to within 0.8 percentage points, against 2.4 before. Because a single pass still samples each task once, we compare post-warmup validation averages rather than individual passes. The full band and its upper half both improve, reaching post-warmup averages of $0.671$ and $0.670$, whereas the lower half reaches $0.640$ and the unscreened pool improves on neither measure, ending at its starting level (Figure~\ref{fig:taskband}). During screening, $72.7\%$ of the unscreened pool's tasks are never solved and $13.4\%$ are always solved, leaving only a small fraction capable of producing within-group reward variation. These results show that difficulty screening is important not merely for task quality, but for maintaining a sufficient density of tasks that provide usable group-relative learning signals.

% \todo{Two sub-studies of the selection pipeline remain unmeasured: a Stage~1 source-pool
% comparison between matched OpenSWE-derived and SWE-rebench-derived subsets, and a Stage~3
% selector-model comparison reporting retained-task overlap and downstream SWE-bench Verified
% performance.}

\FloatBarrier

\subsection{Faithful Optimization}
\label{sec:faithful}

We next evaluate whether training preserves the policy behavior observed during rollout, focusing on token-level probability alignment, routing consistency for sparse models, and generation-time capture under harness-side history rewriting.

\paragraph{Rollout--training alignment.}
% \footnote{\bai{do we define the metrics (briefly) of the alignment? (KL or pearson correlation), and these metrics should be referenced, as they are not invented by us. Explain why we prefer correlation above 0.99?}}
We compare $\pi_{\mathrm{roll}}$, recorded by the proxy at the serving boundary,
with $\pi_{\mathrm{train}}$, recomputed by the trainer over the same assistant
tokens. Across the tokens of a trajectory we report the Pearson correlation
between the two log-probabilities, their mean absolute difference
$|\Delta\log p|$, which is the log per-token importance ratio that enters the
gradient, and the token-averaged
$\mathrm{KL}(\pi_{\mathrm{roll}}\,\|\,\pi_{\mathrm{train}})$ under the $k_3$
estimator \citep{schulman2020kl}; all three are standard checks on
train--inference mismatch \citep{fu2025areal}.
Alignment remains consistently high across OpenHands SDK, Claude Code, and OpenCode: the median Pearson correlation is at least $0.998$ for every scaffold and never falls below $0.989$ at any training step, while the p99 per-trajectory discrepancy in mean token log-probability remains below $3\times10^{-3}$ (Table~\ref{tab:faithful}). These results show that trainer-side probability recomputation closely reproduces rollout-time generation across different agent scaffolds and context-management policies.

\begin{table}[t]
    \centering
    \small
    \setlength{\tabcolsep}{9pt}
    \begin{tabular}{lrrrrr}
        \toprule
        & & & \multicolumn{3}{c}{$|\Delta \overline{\log p}|$ per trajectory ($\times 10^{-3}$)} \\
        \cmidrule(lr){4-6}
        Coding agent & Pearson $r$ & KL ($\times10^{-3}$) & p50 & p90 & p99 \\
        \midrule
        OpenHands SDK & $0.9993$ & $0.75$ & $0.7$ & $1.2$ & $2.1$ \\
        Claude Code   & $0.9980$ & $1.35$ & $0.7$ & $1.3$ & $2.7$ \\
        OpenCode      & $0.9993$ & $0.60$ & $0.6$ & $1.1$ & $2.0$ \\
        \bottomrule
    \end{tabular}
    \caption{Rollout-to-training alignment over the three matched production runs:
    probabilities captured at the serving boundary against trainer-side recomputation over the corresponding assistant tokens. All statistics are medians over training steps.}
    \label{tab:faithful}
\end{table}

\paragraph{Routing consistency.}
For sparse policies, reproducing the same token sequence is insufficient when rollout and training select different experts. Replaying rollout-time routing decisions increases rollout--training correlation from $0.9946$ to $0.9993$ and reduces the mean token log-probability discrepancy from $0.0062$ to $0.0025$ on the same single-node workload. At the first training step, where serving and trainer weights are identical, expert overlap reaches $0.996$ with $0.985$ top-$1$ agreement. Deliberately misaligned replay degrades all alignment measures, confirming that routing decisions must remain associated with the tokens generated under them; additional negative-control results are reported in Appendix~\ref{app:r3}.

\paragraph{Generation-time capture and history consistency.}
Harness-side history processing makes post-hoc trajectory reconstruction unreliable, motivating capture at generation time. In Claude Code production trajectories, tool-call reserialization is the most common apparent mismatch: matching tool calls by identifier rather than serialized arguments resolves $207$ of $222$ ($93\%$) cases without altering the captured tokens. Sub-agent requests may also be mixed with the parent trajectory when they share a session; isolating them into separate sessions eliminates this issue in a subsequent $27$-trial check, from $6.3\%$ to $0\%$. Trajectories interrupted by weight synchronization are retained only when the captured prefix remains exact, whereas histories that are genuinely rewritten or truncated, such as through context compaction, are excluded from optimization. Training is therefore restricted to tokens whose rollout-time identity and probability information can be preserved exactly.

\FloatBarrier

\subsection{Observability and Behavioral Analysis}
\label{sec:behavior}

Trajectory-level observability enables both failure diagnosis and behavioral analysis by linking aggregate training metrics to task-, trajectory-, and execution-level evidence.

\begin{figure}[t]
    \centering
    \begin{minipage}[b]{0.42\linewidth}\includegraphics[width=\linewidth]{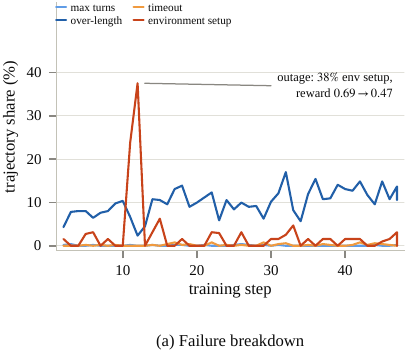}\end{minipage}\hfill
    \begin{minipage}[b]{0.56\linewidth}\includegraphics[width=\linewidth]{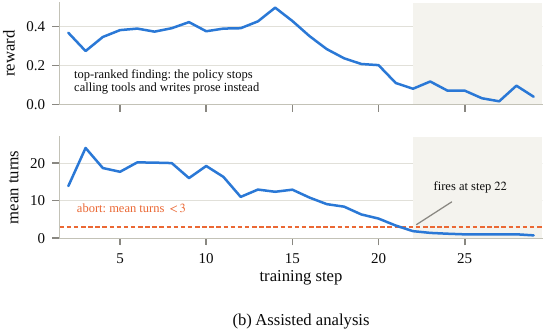}\end{minipage}
    \caption{\textbf{Failure diagnosis with the Live UI.} \textbf{(a)}~Per-step termination reasons for an environment-failure run. \textbf{(b)}~Assisted analysis of a collapsed run and the corresponding early-stop condition. (a) and (b) are two different diagnostic runs, neither is the Claude Code production run reported elsewhere in this section.}
    \label{fig:livediag}
\end{figure}

\paragraph{Failure diagnosis.} The Live UI connects changes in training metrics to the underlying execution and trajectory evidence (Figure~\ref{fig:livediag}). In one run, validation reward fell from $0.556$ to $0.150$, while only $60$ of $172$ trajectories reached verification; termination analysis traced the failure to task setup rather than policy degradation. In another run, all $1{,}024$ trajectories terminated after a single turn, and trajectory inspection identified an incompatible tool-call parser. In a separate collapsed run, trajectory-level analysis revealed the disappearance of valid tool calls and yielded an early-stop condition that would have triggered eight steps before termination. Together, these cases show that the observability layer can distinguish policy degradation from failures in execution, task setup, and agent integration.

\begin{figure}[t]
    \centering
    \begin{minipage}[b]{0.42\linewidth}\includegraphics[width=\linewidth]{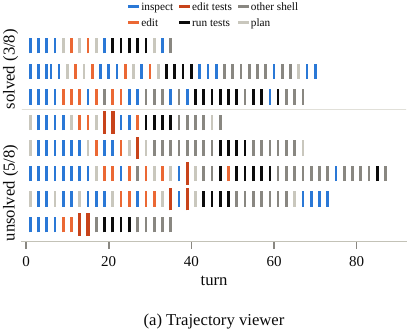}\end{minipage}\hfill
    \begin{minipage}[b]{0.56\linewidth}\includegraphics[width=\linewidth]{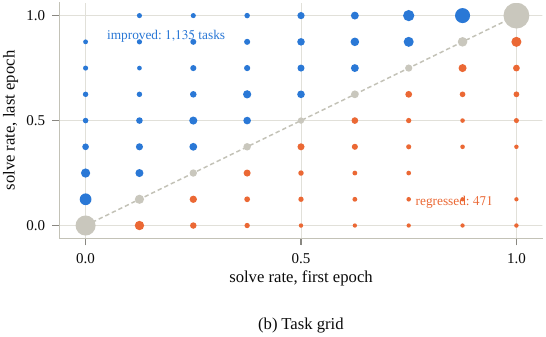}\end{minipage}
    \caption{\textbf{Behavioral analysis with the Live UI, Claude Code run.} \textbf{(a)}~Tool-use trajectories for eight rollouts of one task. \textbf{(b)}~Task-level solve rates in the first and last sampled epochs.
    % \bai{Upper case for the fig subtitle. Check the rest places. BTW, why dont use subfigure?}
    }
    \label{fig:livebehavior}
\end{figure}

\paragraph{Task-level heterogeneity.}
Mean reward improvement can mask substantial variation across individual tasks. The Live UI exposes this variation at task and trajectory granularity (Figure~\ref{fig:livebehavior}). Between the first and last sampled epochs, $1{,}135$ tasks in the Claude Code run increase in observed solve rate while $471$ decrease; OpenHands SDK shows a similar pattern, with $1{,}136$ increasing and $445$ decreasing. The median increase is $0.250$, compared with a median decrease of $0.125$. Because solve rates are estimated from eight asynchronously sampled rollouts per task, these differences are interpreted as observed task-level changes rather than direct measures of learning or forgetting.

\paragraph{Changes in self-checking and error recovery.} In the OpenHands SDK run, agents increasingly inspect their work before and after editing. The proportion of trajectories that reread a file after modifying it rises from $73.6\%$ to $98.1\%$, while the number of distinct files examined before the first edit increases from $3.5$ to $6.9$. By contrast, recovery after intermediate command failures changes only modestly: among affected trajectories, the proportion that ultimately solve the task increases from $63.9\%$ to $66.8\%$. The larger shift in self-checking suggests that training changes verification behavior more strongly than recovery after an error has already occurred. Appendix~\ref{app:behavior} reports the full trajectory-level analysis.

\paragraph{Growth in interaction horizon.}
\label{sec:lengthgrowth}
Response-length growth is driven primarily by more interaction steps rather than longer individual turns. In the OpenHands SDK run, mean response length increases from $43.5$k to $90.9$k tokens, while turns per trajectory increase from $46.6$ to $83.1$ ($+78\%$) and tokens per turn from approximately $933$ to $1{,}093$ ($+17\%$). The increase is smaller under Claude Code, where mean response length grows from $41$k to $51$k under the same model, task pool, and context budget. The realized interaction horizon therefore depends on both the policy and the agent scaffold, with corresponding system-level consequences analyzed in \S\ref{sec:efficiency}. Additional analyses of tool allocation, response composition, and validation failures are reported in Appendices ~\ref{app:toolmix},~\ref{app:cot}, and~\ref{app:valinventory}.

\FloatBarrier

\begin{table}[t!]
    \centering
    \small
    \begin{tabular}{lrrrrr}
        \toprule
        Stage & Mean (s) & p50 & p90 & p99 & Mean-time fraction \\
        \midrule
        Sandbox setup    & $21.6$  & $7.7$   & $41.8$   & $275.2$   & $2.3\%$ \\
        Agent setup      & $4.2$   & $4.0$   & $4.8$    & $8.4$     & $0.5\%$ \\
        Agent execution  & $840.5$ & $708.4$ & $1604.3$ & $2801.1$  & $91.3\%$ \\
        Verification     & $35.9$  & $4.1$   & $29.6$   & $928.5$   & $3.9\%$ \\
        Other            & $20.3$  & $8.1$   & $12.3$   & $965.5$   & $2.2\%$ \\
        \midrule
        Trial total      & $920.4$ & $770.0$ & $1742.7$ & $3189.1$  & $100\%$ \\
        \bottomrule
    \end{tabular}
    \caption{Stage-wise wall-clock statistics across $3{,}699$ OpenHands SDK training trials. Mean-time fraction is computed relative to the mean total trial duration; Other includes scheduling, trajectory handling, and sandbox cleanup.}
    \label{tab:stagetime}
\end{table}

\subsection{System Efficiency}
\label{sec:efficiency}

We evaluate whether \textsc{Lego-RL} can sustain long-horizon coding-agent RL at practical throughput by examining end-to-end runtime, rollout scheduling, sandbox setup, and image delivery.

\paragraph{End-to-end bottleneck and pipeline balance.}
Agent execution dominates the cost of long-horizon coding-agent RL. Across $3{,}699$ OpenHands SDK trials, agent execution accounts for $91.3\%$ of mean wall-clock time, with a mean duration of $840.5$s out of $920.4$s per trial (Table~\ref{tab:stagetime}). Sandbox setup and verification account for only $2.3\%$ and $3.9\%$ on average, although both exhibit substantial tail latency. Under asynchronous training, the median fraction of idle rollout slots is zero for both OpenHands SDK and Claude Code. The trainer nevertheless waits for rollouts for $40.8\%$ and $66.1\%$ of training time, respectively, indicating that rollout generation rather than optimization is the primary throughput bottleneck. This imbalance becomes more pronounced as agent trajectories grow longer.

\begin{figure}[t]
    \centering
    \includegraphics[width=0.68\linewidth]{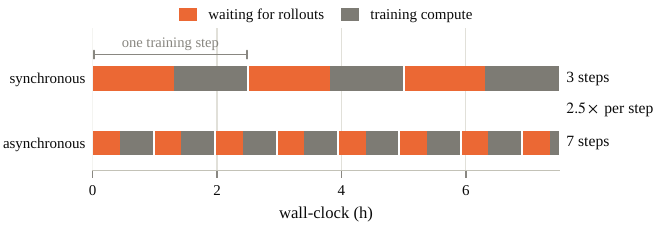}
    \caption{Trainer schedule under synchronous and asynchronous execution.}
    \label{fig:syncasync}
\end{figure}

\paragraph{Asynchronous scheduling.}
Agentic rollouts vary substantially in duration, causing synchronous generation to be delayed by the slowest trajectories. \textsc{Lego-RL} instead schedules rollouts asynchronously, allowing new trajectories to begin without waiting for the preceding batch to complete. In an offline synchronous screening workload using the same agent, sandbox, and verification stack, the slowest $10\%$ of trajectories account for $24.5\%$ of total trajectory work-time. Reconstructed progress timestamps reveal $31$ batch-boundary stalls, with a median duration of $38.7$ minutes and a maximum of $135.9$ minutes. Asynchronous production training avoids these synchronization delays and keeps rollout capacity continuously utilized. The synchronous workload is an offline screening campaign, not a matched training run, so these numbers quantify synchronization overhead and not
end-to-end training speedup.

\paragraph{Synchronous--asynchronous comparison.}
We directly compare synchronous and asynchronous training under matched configurations. Over the same $7.5$ hours, synchronous training completes three steps, whereas asynchronous training completes seven, corresponding to a measured $2.5\times$ reduction in step time (Figure~\ref{fig:syncasync}). The two runs, however, use GPU groups with different optimizer throughput: per trained token, the synchronous run spends $2.1\times$ longer on optimization. After correcting for this compute-rate difference, the estimated synchronous step time decreases from $2.5$ to $1.9$ hours, compared with $1.0$ hour under asynchronous execution. This result is specific to the staleness-$1$ setting used here, while larger staleness may enable greater rollout--optimization overlap.

\begin{figure}[t]
    \centering
    \includegraphics[width=\linewidth]{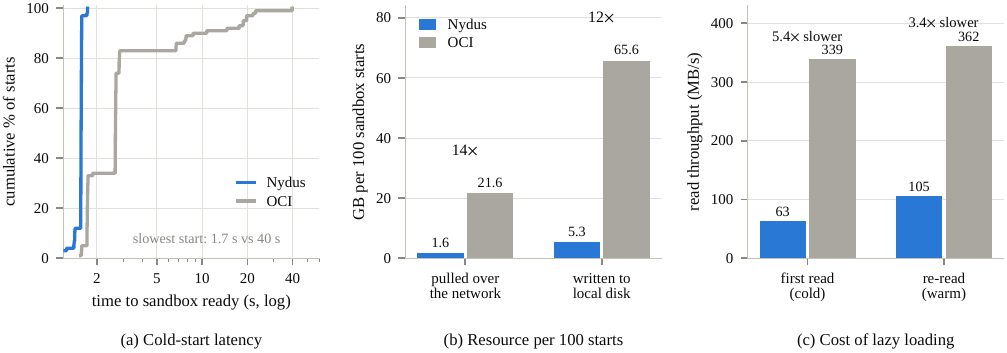}
    \caption{Lazy versus full image delivery over the same $100$ task images.
    Nydus streams image chunks on demand, whereas OCI denotes the conventional
    pull, which materializes the entire image before the container starts. The
    panels report startup latency, cumulative network and disk traffic, and
    in-container read throughput.}
    \label{fig:nydus}
\end{figure}

\paragraph{Sandbox setup.}
Reducing repeated environment construction substantially lowers sandbox startup overhead. Prebuilt task images achieve a median paired speedup of $33.2\times$ over inline Dockerfile execution, while mounting the coding-agent runtime instead of installing it inside each sandbox yields a $15.4\times$ speedup. Table~\ref{tab:ablation} summarizes these ablations together with the image-delivery optimization described below. Packaging grading dependencies, by contrast, provides no measurable latency benefit and is retained primarily to ensure reward reproducibility.

% Per-stage infrastructure ablation. Numbers from ablation_data.json and the
% raw abl_*_data.jsonl; regenerate with collect_ablation.py.
% \input{tab_ablation} ; needs booktabs.
\begin{table}[t]
    \centering
    \small
    \setlength{\tabcolsep}{5pt}
    \begin{tabular}{llccccc}
        \toprule
        & & \multicolumn{2}{c}{Median latency} & \multicolumn{2}{c}{Paired ratio} & \\
        \cmidrule(lr){3-4} \cmidrule(lr){5-6}
        Optimization & Stage & With & Without & Median & p10--p90 & $n$ \\
        \midrule
        Lazy image pull & sandbox setup
          & $1.57$\,s & $2.66$\,s & $1.7\times$ & --- & $100$ \\
        Prebuilt task image & sandbox setup
          & $1.04$\,s & $36.2$\,s & $33.2\times$ & $17.9$--$67.5\times$ & $50$ \\
        Mounted agent runtime & agent setup
          & $0.51$\,s & $7.82$\,s & $15.4\times$ & $14.5$--$16.6\times$ & $50$ \\
        Packaged grading toolchain & verification
          & $3.81$\,s & $2.72$\,s & $0.71\times$ & $0.67$--$0.73\times$ & $50$ \\
        \bottomrule
    \end{tabular}
    \caption{Ablation of sandbox optimizations.}
    \label{tab:ablation}
\end{table}

\paragraph{Lazy image delivery.}
Lazy image delivery reduces the remaining startup cost by fetching image data on demand rather than pulling the full task image before execution. Across $100$ SWE-bench Verified task images, lazy pull improves median cold-start latency by $1.7\times$ and maximum latency by $23\times$, while reducing cumulative network traffic from $21.6$\,GB to $1.59$\,GB and disk writes from $65.6$\,GB to $5.29$\,GB (Figure~\ref{fig:nydus}). The gains are largest when coding-agent workloads access only a small fraction of the image and disappear on warm starts, where both delivery paths perform similarly. The trade-off is lower uncached in-container read throughput because missing blocks are fetched remotely. In production, we additionally use a shared snapshotter daemon to prevent the daemon accumulation observed with the default per-image mode, trading per-image isolation for lower long-term resource overhead.

\FloatBarrier

\section{Conclusion}
\label{sec:conclusion}

We present \textsc{Lego-RL}, a framework that connects native coding-agent harnesses to scalable policy-gradient optimization while preserving their original control flows. By integrating sandboxed execution and verification, token-faithful rollout capture, asynchronous training, reward-integrity safeguards, and trajectory-level observability, \textsc{Lego-RL} supports reliable and faithful training across OpenHands SDK, Claude Code, and OpenCode.
% Our results also suggest two directions for scaling coding-agent RL: expanding toward larger and more adaptive task pools as policy-relative task difficulty evolves during training, and developing richer credit assignment beyond terminal binary rewards to better reinforce behaviors such as error recovery.
Together, these results highlight that scaling coding-agent RL requires not only scalable optimization, but also reliable execution environments, faithful trajectory capture, and feedback mechanisms that evolve with increasingly capable agents.

\section*{Limitations and Future Work}
\label{sec:limitations}

% \footnote{\bai{Briefly mention future plans? e.g., We will continuously update the released framework, inlcuding 1) support mulitiple swe-task joint training; 2) multi-harness joint training; 3) etc.. }}

\paragraph{Limitations.}
Our evaluation has several limitations. First, all experiments use Qwen3.5-35B-A3B, and each coding-agent harness is trained separately, so generalization to other model architectures and mixed-harness training remains to be evaluated. Second, the cost of production-scale training limits each main configuration to a single run, leaving run-to-run variance in training gains and system-level efficiency unquantified. Third, executable verification provides a reliable but coarse binary reward and cannot assign intermediate credit to behaviors such as error recovery. Our reward-integrity defenses address the failure modes observed in our experiments but do not guarantee robustness to all possible reward-exploiting strategies. Finally, the reported sandbox and image-delivery speedups depend on the deployment environment and should be interpreted as measurements of our implementation rather than universal properties of \textsc{Lego-RL}. More broadly, the diagnostic analyses presented here identify plausible causes from execution and trajectory evidence, but do not constitute automated causal verification.

\paragraph{Future work.}
Several of these limitations point at work already under way. We are extending \textsc{Lego-RL} to mixed training: one policy over a task pool that mixes repository repair with other verifiable software tasks, and one policy trained
across several harnesses at once rather than one run per harness---the latter motivated by the harness dependence visible in Table~\ref{tab:swe_bench_comparison}. We are also adding adapters for further harnesses, richer credit assignment than terminal binary rewards, and automated diagnosis in the Live UI. \textsc{Lego-RL} is developed in the open; we will continue to release framework updates, harness adapters, trained checkpoints, and task indices.

\bibliographystyle{plainnat}
\bibliography{references}

\clearpage
\appendix

\section{Reward-Integrity Failure Modes}
\label{app:antihack}

The reliability of our training pipeline depends on the verifier reward faithfully reflecting task completion. During development, we identified six systematic failure modes that violate this assumption. Table~\ref{tab:antihack} categorizes these failures by their root cause---whether the agent exploits the reward mechanism, or the environment produces a reward decoupled from agent behavior---and lists the corresponding defenses deployed in production.

\begin{table}[h]
    \centering
    \small
    \setlength{\tabcolsep}{10pt}
        \begin{tabular}{p{0.28\linewidth} c p{0.45\linewidth}}
        \toprule
        Failure Mode & Incidence & Defense \\
        \midrule
        \multicolumn{3}{l}{\textit{Agent-side: shortcut exploitation}} \\
        Reads git history & $4.6$--$20.5\%$ & Rebase history to a single commit during agent phase; restore before grading \\
        Downloads reference fix & $1.9\%$ & Per-phase egress firewall in privilege-separated sidecar \\
        Edits test files & $2.4$--$19.4\%$ & Withhold tests until grading; revert test-path edits \\
        \midrule
        \multicolumn{3}{l}{\textit{Environment-side: reward detached from agent}} \\
        Grader applies reference patch & $2.5\%$ & Audit affected instances out of pool; flag live if degenerate reward propagates \\
        Grader requires network access & --- & Package all grade-time dependencies; ensure deterministic invocation \\
        Incomplete repository build & --- & Hermetic fail-fast build; report setup failure explicitly, not as zero reward \\
        \bottomrule
    \end{tabular}
    \caption{Reward-integrity failure modes and their mitigations. Incidence rates are measured prior to deploying the defenses; ``---'' indicates cases not separately quantified.}
    \label{tab:antihack}
\end{table}

\section{Routing-Replay Negative Control and Capture Coverage}
\label{app:r3}

The routing-replay mechanism introduced in \S\ref{sec:faithful} is subject to two independent failure modes: incorrect alignment whether a replayed routing decision is assigned to the correct token. The second concerns completeness: whether a routing decision was recorded at all. These are independent failures—a system can have perfect alignment but incomplete capture, or complete capture but systematic misalignment.

\paragraph{Misaligned replay.}
An intermediate implementation introduced a systematic one-position offset between each token and its replayed routing decision, routing every token through the experts selected for its neighbor. Table~\ref{tab:r3app} shows that this misalignment degrades all metrics relative to disabling replay entirely. Crucially, this degradation is not detectable by merely verifying that the replay mechanism is active—the system appears to function normally. The only diagnostic that separates correct from incorrect replay is a direct comparison of the replayed expert assignments against the model's own online selections, which is why we report expert overlap and top-1 agreement in the Table~\ref{tab:r3app}.

\begin{table}[h]
    \centering
    \small
    \setlength{\tabcolsep}{6pt}
    \begin{tabular}{lcccc}
        \toprule
        Routing replay configuration & Pearson $r$ & Mean $|\Delta p|$ & Expert overlap & Top-1 agreement \\
        \midrule
        Disabled            & $0.9946$ & $0.0062$ & ---     & ---     \\
        Enabled, misaligned & $0.7503$ & $0.0954$ & $0.083$ & $0.026$ \\
        Enabled, aligned    & $0.9993$ & $0.0025$ & $0.996$ & $0.985$ \\
        \bottomrule
    \end{tabular}
    \caption{Routing-replay configurations compared. Expert overlap and top-1 agreement are undefined when replay is disabled.}
    \label{tab:r3app}
\end{table}

\paragraph{Incomplete capture.}
For replay to be complete, routing decisions must be recorded for every generated token. Our initial capture buffer, however, was sized using a formula that underestimated the requirement for hybrid-attention models by roughly a factor of four. Worse, the out-of-range guard recorded excess decisions as zero rather than raising a failure, making the problem silent. Coverage therefore decayed with sequence length, reaching only $24\%$ overall before diagnosis. After resizing the buffer, both production runs now exceed $99.8\%$ coverage. The residual misses occur only on the longest sequences and degrade gracefully: tokens without a recorded decision are replayed unconstrained, rather than being forced through an incorrect assignment. This fail-soft behavior ensures that incomplete capture, unlike misalignment, does not actively harm the policy.

\section{Assisted-Analysis Case Study: A Collapsed Run}
\label{app:analysiscase}

The run summarized in Figure~\ref{fig:livediag}(b) serves as an end-to-end illustration of the assisted-analysis pane. The configuration is Qwen3-30B-A3B with the OpenHands SDK on a $449$-task pool; training reward fell from $0.351$ to $0.050$ and held-out validation reward from $0.230$ to $0.014$ over $29$ logged steps. The text that follows is the pane's output for this run, edited only for formatting; each claim was verified against the raw trajectories before inclusion. The pane is required to mark missing data sections as such rather than infer them; for this run, three sections—zero-advantage accounting, first-versus-last trajectory shape, and per-task solve statistics—were not populated. Table~\ref{tab:analysiscase} reports the numeric evidence supplied to the pane: the mean of the first five logged steps against the mean of the last five, with a least-squares trend statistic ($t$) computed over the entire run.

\begin{table}[h]
    \centering
    \small
    \begin{tabular}{lrrr}
        \toprule
        Series & First five steps & Last five steps & $t$ \\
        \midrule
        training reward              & $0.351$   & $0.050$   & $-7.5$ \\
        turns per trajectory         & $18.9$    & $0.96$    & $-14.8$ \\
        response length (tokens)     & $12{,}170$ & $2{,}276$ & $-7.7$ \\
        tokens per turn              & $650$     & $2{,}397$ & --- \\
        policy entropy               & $0.134$   & $0.230$   & $+7.2$ \\
        KL term of the loss          & $0.0046$  & $0.130$   & $+3.4$ \\
        rollout--training agreement  & $0.995$   & $0.970$   & $-3.3$ \\
        gradient norm                & $0.31$    & $0.034$   & $+0.4$ \\
        \bottomrule
    \end{tabular}
    \caption{Evidence supplied for the collapsed run. $t$ is the least-squares slope divided by
    its standard error over all logged steps; the gradient norm does not clear $|t|=2$ and is
    therefore read as noise. Tokens per turn is a ratio of two rows above it and carries no
    separate trend statistic.}
    \label{tab:analysiscase}
\end{table}

The pane identifies step~17 as the point where learning ceased. The training reward decline ($t=-7.5$) is mirrored by validation reward, ruling out a training-metric artifact. The proximate failure is the policy's cessation of action: between steps~17 and~25, turns per trajectory fell from $9.0$ to $1.0$ and did not recover, response length shrank to one-fifth of its prior value, and reward dropped to a tenth. Trajectory inspection confirms the mechanism: a representative late trajectory contains a single turn with no tool call, with intended shell commands appearing as fenced code blocks in the model's prose. Rising entropy ($t=+7.2$) over the same interval is consistent with degeneration, not exploration.

The group-level advantage structure collapses as nearly every group becomes all-wrong: advantage bounds reach zero at two of the last three steps, while the KL term rises to $150\times$ its initial level, leaving the update dominated by penalty rather than reward. Rollout--training agreement declines from $0.995$ to $0.851$ at step~22. The evidence does not distinguish whether this capture regression contributed to the collapse or merely reflects a policy far from its reference.

Infrastructure failures do not explain the collapse: setup-failure share never exceeded $1.6\%$ of rollouts across steps~17--28, and thus postdates the collapse by more than ten steps. The case yields several monitoring criteria: a mean turn count approaching one is terminal; a batch with no partially solved group reached $100\%$ at two of the last three steps; and rollout--training agreement crossed below $0.95$ at step~21, eight steps before manual termination.

\section{How Validation Attempts Fail}
\label{app:valinventory}

A verifier reward records only whether a validation attempt succeeded. To see what changes
among the attempts that do not, we classify every trajectory of a validation event by how far
the attempt got: whether it edited the right file, edited it without fixing the issue, produced
no edit, or exhausted its budget. Over the
OpenHands SDK production run, comparing the first and last of its $26$ validation events on the
same $500$-task set, the resolved share rises from $63.8\%$ to $68.6\%$ and the remainder
redistributes. Attempts that edit the wrong file halve, from $6.4\%$ to $3.0\%$ of the set,
and attempts that edit the file the reference patch edits without resolving the issue fall
from $26.2\%$ to $21.8\%$. One category moves the other way, from $1.6\%$ to $5.4\%$:
trajectories that exhaust their budget in a repeated pattern without progress. Attempts that
produce no edit at all fall from $1.6\%$ to $0.8\%$, and environment or verifier failures hold
at $0.4\%$.
Over training, localization failures decrease, patch-quality failures remain
dominant, and budget-exhaustion failures become more frequent---the same shift
\S\ref{sec:lengthgrowth} reports on the training set, here visible in the validation
set.
% Over training, localization failures decrease, patch-quality failures remain dominant, and budget-exhaustion failures become more frequent, as shown in \S\ref{sec:lengthgrowth} appearing in the validation set.
The classification is rule-based rather than a model judgment, and its categories are not equally
reliable: the localization and no-edit categories are read from the actions taken, whereas
budget exhaustion is inferred from repetition.

\section{Reasoning–Action Composition Across Training}
\label{app:cot}

\begin{figure}[t!]
    \centering
    \includegraphics[width=\linewidth]{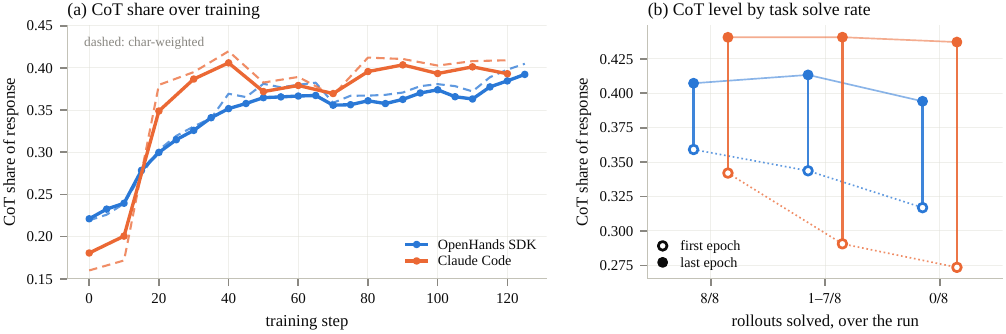}
    \caption{Reasoning share of the response over training on a fixed $120$-task validation cohort. (a) Mean over tasks (solid) and character-weighted mean (dashed); (b) Median share at each task's first (open) and last (filled) sampled epoch, grouped by number of rollouts solved.}

    \label{fig:behavior}
\end{figure}

We define the reasoning share of a response as the proportion of characters not contained within tool-call delimiters. Tool-call arguments are classified as acting rather than reasoning, since they correspond to the agent's executable modifications to the codebase. (Scoring against visible prose alone would roughly double every level reported here.) The metric is computed on a fixed $120$-task cohort subsampled from the SWE-bench Verified validation set and reused at every validation event of both runs, ensuring comparability across steps. Figure~\ref{fig:behavior}(a) reports both the per-task mean and the character-weighted mean; the two agree closely throughout, indicating that long trajectories neither dominate the average nor are discounted by it. The Claude Code curve in Figure~\ref{fig:behavior}(a) begins at a step-0 point from an earlier run of the identical configuration, as the production run's own step-0 transcripts were not retained; the validation reward at that step matches the production run's step-0 value.

Figure~\ref{fig:behavior}(a) shows that reasoning share rises over training under both scaffolds, from $0.22$ to $0.39$ (OpenHands SDK) and from $0.18$ to $0.39$ (Claude Code). The two scaffolds start apart but converge to the same level, pointing to a policy-level trend rather than a scaffold effect. This interpretation is observational: a shared training objective, task pool, and base model could equally produce the same pattern, and the measurement does not isolate a cause.

Panel~(b) examines whether the increase is concentrated on tasks the run finds difficult. Grouping tasks by how many of their eight rollouts the run solved, the early-training reasoning share decreases with task difficulty; by the last epoch, all three groups sit at nearly the same level. Hard tasks gain the most simply because they start the lowest, not because reasoning grows without bound where it is needed. Since the grouping is based on post-training outcomes, this does not establish that additional reasoning causes success---that would require direct intervention on reasoning length. We therefore interpret the result as a drift in trajectory shape toward a common reasoning ratio, not as evidence of improved reasoning capability.

\section{Tool Allocation}
\label{app:toolmix}

\begin{figure}[t!]
    \centering
    \includegraphics[width=\linewidth]{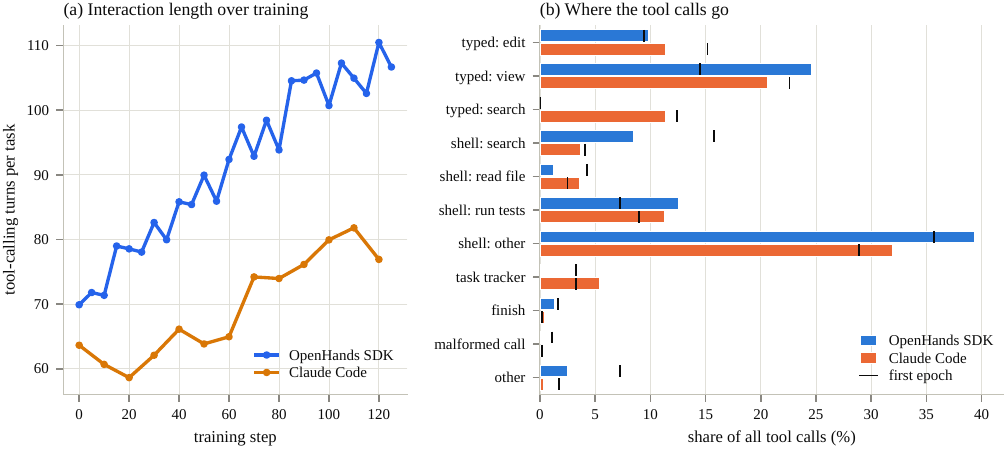}
    \caption{Agent behaviors before and after training, computed over $420$ trajectories at each end of the production OpenHands SDK run. The pass@$k$ and pass$^k$ rows are computed over prompt groups in the first and last third of the run.}

    \label{fig:capability}
\end{figure}

The interaction horizon grows under both scaffolds: tool-calling turns per validation task rise from $69.9$ to $106.7$ (OpenHands SDK) and from $63.6$ to $76.9$ (Claude Code), as shown in Figure~\ref{fig:capability}(a). Tool allocation shifts in scaffold-specific directions, but two trends are shared: test-suite invocation becomes more frequent and malformed calls decline. On a common file-operation axis, the initial difference between the two scaffolds narrows from $13.9$ percentage points ($23.9\%$ vs.\ $37.8\%$) to $2.4$ points ($34.4\%$ vs.\ $32.0\%$).

Panel~(b) of Figure~\ref{fig:capability} provides the per-category breakdown over the first and last $420$ trajectories of each run. OpenHands SDK reduces shell-based file inspection from $20.1\%$ to $9.7\%$ of calls while its structured view command rises from $14.5\%$ to $24.6\%$; Claude Code moves in the opposite direction, with structured file tools falling from $37.8\%$ to $32.0\%$ and shell inspection rising from $6.5\%$ to $7.3\%$. This comparison is approximate, however, as the two harnesses expose different tool sets; only Claude Code offers a typed search, so OpenHands SDK performs the equivalent work via shell commands. We therefore do not claim convergence of tool behavior in general.

Figure~\ref{fig:capability} thus captures the aggregate tool-use dynamics. The finer-grained agent behaviors that underlie these aggregates are presented in Table~\ref{tab:capability} below.

\section{Agent Behavior Changes Across Training}
\label{app:behavior}

Table~\ref{tab:capability} is the full set of trajectory-level behaviors behind
\S\ref{sec:behavior}, each read off the agent transcript by a deterministic scan rather than
by a judge. The upper block is self-verification and exploration, which improve
substantially; the middle block is failure handling, which barely moves; the lower blocks
give protocol compliance and the coverage/reliability split discussed in the main text.

\begin{table}[h]
    \centering
    \small
    \begin{tabular}{lrrr}
        \toprule
        Behavior & First & Last & $\Delta$ \\
        \midrule
        Reads back a file it edited     & $73.6\%$ & $98.1\%$ & $+24.5$ \\
        Runs the test suite             & $85.0\%$ & $93.6\%$ & $+8.6$ \\
        Files explored before 1st edit  & $3.45$   & $6.92$   & $+3.47$ \\
        Ends with an explicit finish    & $88.3\%$ & $91.9\%$ & $+3.6$ \\
        \midrule
        Reproduces failure before editing & $6.7\%$  & $11.2\%$ & $+4.5$ \\
        Solves despite a failed command   & $63.9\%$ & $66.8\%$ & $+2.9$ \\
        \midrule
        Malformed tool calls (of all calls) & $1.07\%$ & $0.15\%$ & $-0.92$ \\
        \midrule
        Coverage, pass@$8$              & $83.2\%$ & $87.9\%$ & $+4.7$ \\
        Reliability, pass$^8$           & $28.3\%$ & $39.4\%$ & $+11.1$ \\
        \bottomrule
    \end{tabular}
    \caption{Agent behaviors before and after training, over $420$ trajectories at each
    end of the production OpenHands SDK run; the pass@$k$ rows are over prompt groups in the
    first and last third of the run.}
    \label{tab:capability}
\end{table}

\section{Live Observability Dashboard}
\label{app:ui}

\begin{figure}[t!]
    \centering
    \includegraphics[width=\linewidth]{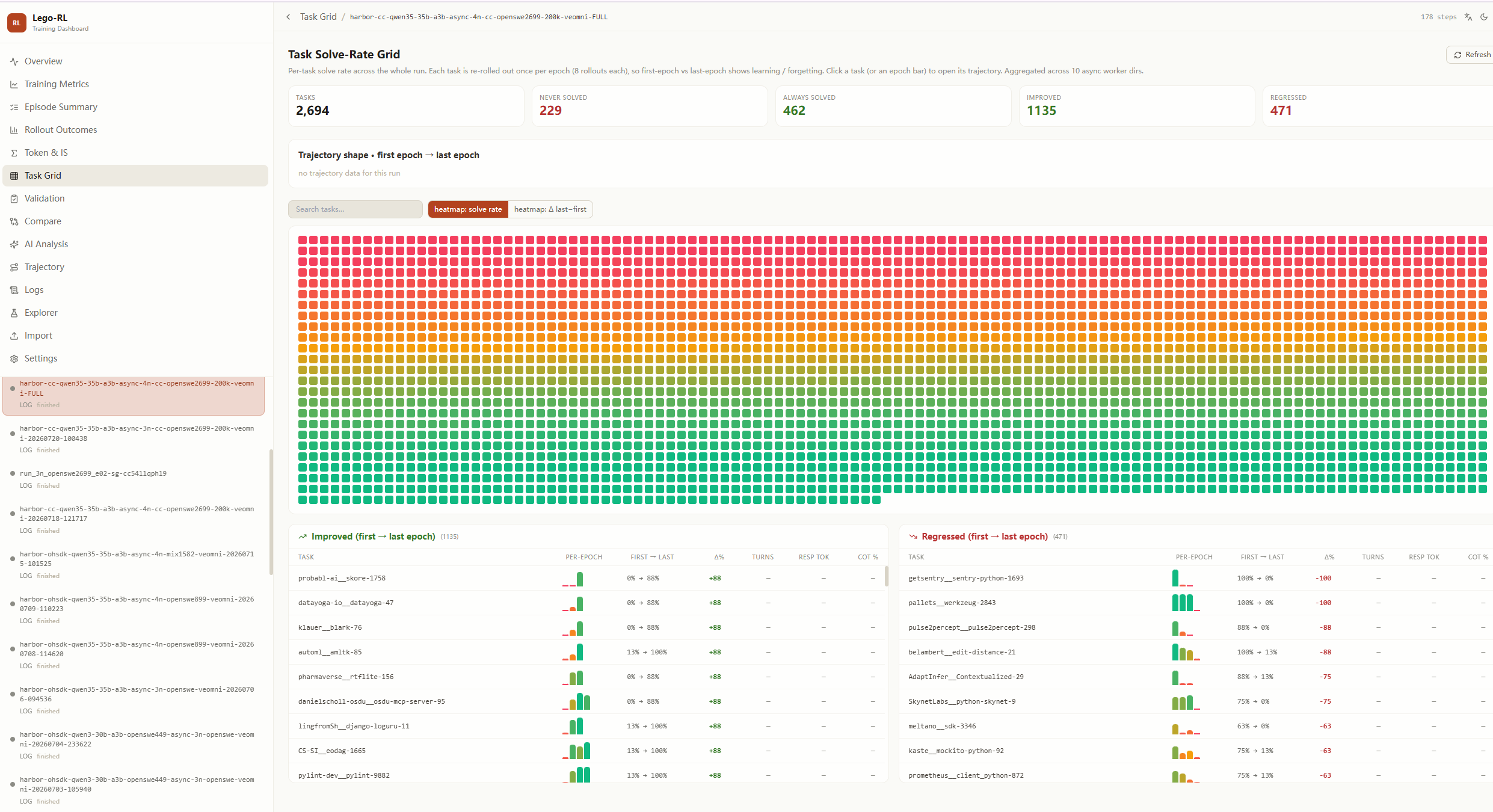}
    \caption{The per-instance task grid.}
    \label{fig:ui-taskgrid}
\end{figure}

\begin{figure}[t!]
    \centering
    \includegraphics[width=0.92\linewidth]{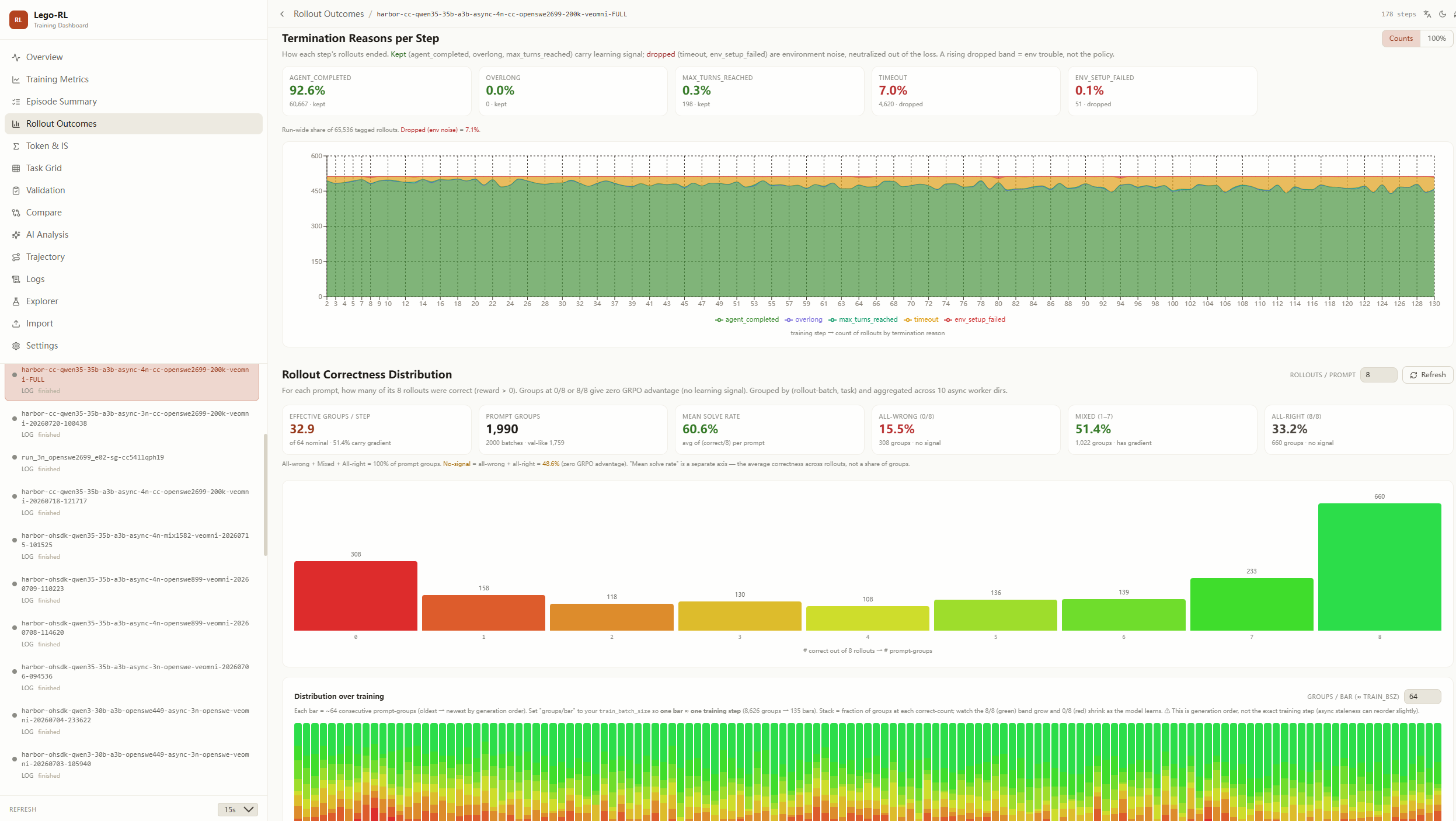}\\[0.5em]
    \includegraphics[width=0.92\linewidth]{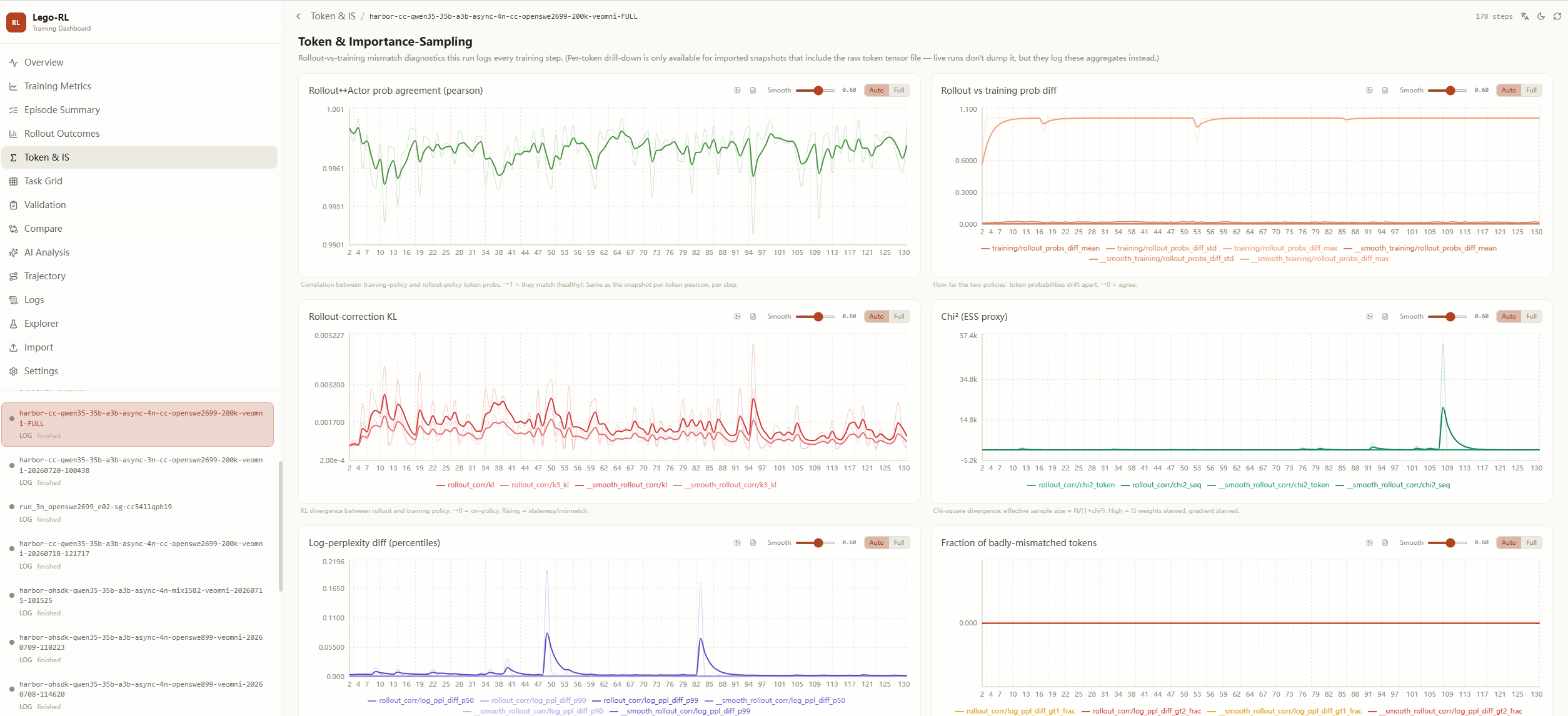}
    \caption{Diagnostic panels from the same run. \emph{Top:} termination-reason breakdown
    per step. \emph{Bottom:} the train--inference consistency panel.}
    \label{fig:ui-diag}
\end{figure}

Figures~\ref{fig:ui-taskgrid} and~\ref{fig:ui-diag} show the observability UI of
\S\ref{sec:dashboard} on the Claude Code run analyzed in
Figure~\ref{fig:behavior}, read while that run was still in flight (Qwen3.5-35B-A3B,
$200$k-token context budget, step $103$); its shares therefore differ from the whole-run
figures of \S\ref{sec:data}.

In the per-instance task grid, each cell is one training task colored by its solve rate
across the run; the header decomposes the aggregate reward into improved versus regressed
tasks (first epoch vs.\ last), and the trajectory-shape strip tracks turn counts, token
budgets, and the chain-of-thought share of responses as training proceeds. The
termination-reason breakdown shows that $94.1\%$ of $34{,}816$ rollouts complete and carry
learning signal, while timeouts and environment-setup failures are classified as environment
noise and neutralized out of the loss.

\paragraph{Panel inventory.} Beyond standard RL curves (entropy, KL, gradient norm, throughput,
model-flops utilization), the dashboard provides:

\begin{itemize}[leftmargin=1.4em,itemsep=2pt,topsep=3pt]
  \item \textbf{Trajectory viewer.} Every trial's full agent transcript rendered next to its
  verifier reward and termination reason, so a suspicious curve can be traced to concrete behavior.
  \item \textbf{Per-instance task grid.} Which task instances flip from unsolved to solved
  (or regress) across checkpoints, as an instance $\times$ checkpoint grid with a trend
  significance test.
  \item \textbf{Consistency panel.} The per-step fidelity suite of \S\ref{sec:faithful}:
  probability Pearson, per-sample mean/max absolute log-ratio, the batch log-ratio histogram,
  and the ESS fraction under importance weighting.
  \item \textbf{In-batch distribution.} The share of the batch whose groups are partially
  solved and thus drive a group-relative update, plus the overlong ratio and length
  distribution.
  \item \textbf{Failure breakdown.} The per-step termination-reason mix together with
  per-stage timing and per-tool frequency---the first panels consulted when reward drops,
  since they separate infrastructure incidents from learning dynamics.
  \item \textbf{Validation failure inventory.} Each trajectory of a validation event
  classified by how far the attempt got, which turns a validation reward into a direction
  (Appendix~\ref{app:valinventory}).
  \item \textbf{AI-assisted analysis.} An optional pane that summarizes a run's metrics and
  trajectories with a language model acting as an RL diagnostician, surfacing candidate
  explanations for a reviewer to confirm (Appendix~\ref{app:analysiscase} works one case end
  to end).
\end{itemize}

Three supporting mechanisms make these views trustworthy during a live run. A central per-step
\emph{progress aggregator} collects a one-line summary from every finished trial (task,
termination reason, reward, turns, wall-clock) and renders a live count with per-outcome
tallies (``$234/512$ trials, $12$ setup failures''), so an operator sees a step forming rather
than waiting for its aggregate. When validation is suspected of scoring \emph{false zeros} (a
verifier artifact rather than a policy failure), an \emph{offline regrade} path re-scores the
affected trials outside the training loop and reconciles the curve, and infrastructure-failed
trials can be purged from an index before they contaminate later analysis. Finally, any run view
can be exported as a \emph{static snapshot} and published as a standalone page, so a training
incident can be shared and inspected without access to the cluster.

\clearpage
\section{Run Configuration}
\label{app:hparams}

Table~\ref{tab:hparams} lists the hyperparameters shared by the three production runs reported in \S\ref{sec:effectiveness}. Four entries require clarification. First, the KL term enters the loss only (never the reward) and uses the low-variance estimator. Second, the context budget is a split limit: prompt and response are truncated independently when either exceeds its respective share. Third, importance-sampling correction is disabled in all three runs because measured fidelity remained high throughout (\S\ref{sec:faithful}); it is enabled only when log-probability diagnostics indicate drift. Fourth, the staleness threshold of $1$ permits full asynchrony while bounding the maximum policy lag of any consumed rollout.

\begin{table}[h]
    \centering
    \small
    \begin{tabular}{ll}
        \toprule
        Policy loss & GSPO \\
        Clip range (sequence-level) & $(3\times10^{-4},\,4\times10^{-4})$ \\
        Advantage estimator & GRPO \\
        Loss aggregation & seq-mean-token-mean \\
        KL reward penalty & none \\
        KL loss coefficient & $10^{-3}$ \\
        \midrule
        Learning rate & $1\times10^{-6}$ \\
        Learning-rate schedule & constant \\
        Gradient clip & $1.0$ \\
        Prompts per batch & $64$ \\
        Rollouts per prompt & $8$ \\
        Micro-batch per GPU & $1$ \\
        \midrule
        Rollout temperature & $1.0$ \\
        Rollout top-$p$ & $1.0$ \\
        Validation temperature & $0.7$ \\
        Validation samples per instance & $1$ \\
        \midrule
        Prompt budget & $30$k tokens \\
        Response budget & $170$k tokens \\
        \midrule
        Staleness threshold & $1$ \\
        Partial-rollout recovery & on \\
        Importance-sampling correction & off \\
        \midrule
        Training pool & $2{,}699$ tasks \\
        Epochs & $3$ \\
        \bottomrule
    \end{tabular}
    \caption{Resolved hyperparameters of the three production runs (Qwen3.5-35B-A3B through
    the OpenHands SDK, Claude Code, and OpenCode).}
    \label{tab:hparams}
\end{table}

\section{Data Formats}
\label{app:data}

This appendix documents the three data representations a task passes through, using the real
instance \texttt{12rambau\_\_sepal\_ui-814} from the OpenSWE-derived pool (long fields
truncated).

\paragraph{Raw instance.}
Each upstream instance is a SWE-bench-style record: a repository snapshot, the issue text, the
gold patch, the test patch, and the test lists that define the verifier outcome.

\begin{figure}[H]
\begin{lstlisting}[style=datafmtjson]
{
  "repo": "12rambau/sepal_ui",
  "instance_id": "12rambau__sepal_ui-814",
  "base_commit": "6d825ae167f96ad2e7b76b96ca07de562f74dcf0",
  "patch": "diff --git a/sepal_ui/sepalwidgets/alert.py ...   (truncated)",
  "test_patch": "diff --git a/tests/test_sepalwidgets/test_Alert.py ...   (truncated)",
  "problem_statement": "avoid to force developer to set total each time\n
      I should be able to init the progress of an Alert first and then
      simply update the progress. ...   (truncated)",
  "FAIL_TO_PASS": ["tests/test_sepalwidgets/test_Alert.py::test_update_progress"],
  "PASS_TO_PASS": ["tests/test_sepalwidgets/test_Alert.py::test_init", ...],
  "environment_setup_commit": "b91b2a2c45b4fa80a7a0c699df978ebc46682260",
  "docker_image": "sweb.eval.x86_64.12rambau_1776_sepal_ui-814:latest",
  "install_config": {"install": "pip install -e .[dev]",
                     "log_parser": "parse_log_pytest", ...}
}
\end{lstlisting}
\caption{Raw instance record for \texttt{12rambau\_\_sepal\_ui-814}.}
\label{fig:fmt-raw}
\end{figure}

\paragraph{Harbor task.}
Data preparation converts each raw instance into an executable Harbor task: a self-contained
directory whose manifest declares the environment image and per-stage resource and timeout
budgets, and whose verifier reproduces the official SWE-bench grading inside the sandbox.

\begin{figure}[H]
\begin{lstlisting}[style=datafmttree]
12rambau__sepal_ui-814/
|-- task.toml                # manifest: image, resources, per-stage timeouts
|-- instruction.md           # the issue text shown to the agent
|-- environment/Dockerfile   # FROM <task image>; WORKDIR /testbed; ...
`-- tests/
    |-- test.sh              # verifier: reset tests -> apply test_patch ->
    |                        #   run -> grade (FAIL_TO_PASS pass, no regression)
    |-- test.patch           # the held-out test patch
    |-- parser.py            # output parser
    `-- config.json          # the raw instance record above

# task.toml (excerpt)
[environment]
docker_image = "sweb.eval.x86_64.12rambau_1776_sepal_ui-814:latest"
cpus = 1
memory_mb = 4096
build_timeout_sec = 1800.0
[agent]
timeout_sec = 3000.0
[verifier]
timeout_sec = 3000.0
\end{lstlisting}
\caption{The same instance as an executable Harbor task: directory layout and a manifest
excerpt.}
\label{fig:fmt-task}
\end{figure}

\paragraph{Agent-visible vs.\ verifier-only state.}
The directory layout in Figure~\ref{fig:fmt-task} is host-side task state, not the agent's filesystem. The agent phase (environment setup, agent execution) sees only the initialized repository. The \texttt{tests/} directory---containing \texttt{test.patch}, \texttt{config.json} (with the gold patch and \texttt{FAIL\_TO\_PASS}/\texttt{PASS\_TO\_PASS} lists), and grading scripts---is uploaded into the sandbox only during the verifier phase, after the agent has finished. This implements the ``tests withheld until grading'' defense described in Table~\ref{tab:antihack}.

\paragraph{Task-index row.}
The trainer samples from a thin task index; each row only points at a Harbor task, so the task
content stays on shared storage and the index stays cheap to filter and re-mix.

\begin{figure}[H]
\begin{lstlisting}[style=datafmtjson]
{ "prompt": [{"role": "user", "content": "<path>/openswe_filtered/..."}],
  "reward_model": {"style": "rule", "ground_truth": null},
  "extra_info": {"data_source": "harbor",
                 "harbor_task_path": "<path>/openswe_filtered/...",
                 "instance_id": "4Catalyzer__flask-annex-37"} }
\end{lstlisting}
\caption{A task-index row: a pointer to a Harbor task rather than the task itself.}
\label{fig:fmt-index}
\end{figure}

\end{document}